\documentclass[10pt,twocolumn,letterpaper]{article}

\usepackage[pagenumbers]{cvpr} 

\usepackage{graphicx}
\usepackage{amsmath}
\usepackage{amssymb}
\usepackage{booktabs}
\usepackage{bbm}

\usepackage[pagebackref,breaklinks,colorlinks]{hyperref}

\usepackage[capitalize]{cleveref}
\crefname{section}{Sec.}{Secs.}
\Crefname{section}{Section}{Sections}
\Crefname{table}{Table}{Tables}
\crefname{table}{Tab.}{Tabs.}

\def\cvprPaperID{2157} 
\def\confName{CVPR}
\def\confYear{2023}

\usepackage[inline]{enumitem} 
\usepackage{color}
\usepackage{multirow}
\usepackage{cases}
\usepackage[makeroom]{cancel}
\usepackage[accsupp]{axessibility}

\usepackage{caption}
\definecolor{turquoise}{cmyk}{0.65,0,0.1,0.3}
\definecolor{purple}{rgb}{0.65,0,0.65}
\definecolor{dark_purple}{rgb}{0.5,0,0.5}
\definecolor{dark_green}{rgb}{0, 0.5, 0}
\definecolor{orange}{rgb}{0.8, 0.6, 0.2}
\definecolor{red}{rgb}{0.8, 0.2, 0.2}
\definecolor{darkred}{rgb}{0.6, 0.1, 0.05}
\definecolor{blueish}{rgb}{0.0, 0.3, .6}
\definecolor{light_gray}{rgb}{0.7, 0.7, .7}
\definecolor{pink}{rgb}{1, 0, 1}
\definecolor{greyblue}{rgb}{0.25, 0.25, 1}

\newcommand{\Figure}[1]{Figure~\ref{fig:#1}}

\newcommand{\Table}[1]{Table~\ref{table:#1}}
\newcommand{\eq}[1]{\eqref{eq:#1}}

\newcommand{\Section}[1]{Section~\ref{sec:#1}}

\usepackage{blindtext}

\renewcommand{\paragraph}[1]{\vspace{\parskip}\noindent\textbf{#1}.}

\newcommand{\SupplementaryMaterial}{{{\color{dark_purple}supplementary}}\xspace}

\usepackage{enumitem}
\setlist[itemize]{noitemsep,leftmargin=*,topsep=0in}
\setlist[enumerate]{noitemsep,leftmargin=*,topsep=0in}

\newcommand{\floatersfootnote}{\footnote{For real-world scenes, we further incorporate the distortion loss $\mathcal{L}_{dist}$ introduced by \cite[Eq.~15]{barron2022mipnerf360} to suppress \textit{floaters} and \textit{background collapse}.}}
\newcommand{\upperboundfootnote}{\footnote{This loss performs a \textit{stochastic upper-bound}, as we initialize $\Grid[*]{=}\mathbf{0}$, and $\Grid[\mathbf{p}_k]$ receives gradients whenever $T_k \alpha_k {>} \Grid[\mathbf{p}_k]$.}}

\DeclareMathOperator*{\argmin}{arg\,min}

\newcommand{\real}{\mathbb{R}}

\newcommand{\Mesh}{\mathcal{M}}
\newcommand{\Triangles}{\mathcal{T}}
\newcommand{\Vertices}{\mathcal{V}}
\newcommand{\expect}{\mathbb{E}}
\newcommand{\ray}{\mathbf{r}}
\newcommand{\pars}{\theta}
\newcommand{\point}{\mathbf{p}}
\newcommand{\opacity}{\alpha}
\newcommand{\dOpacity}{\hat{\alpha}}
\newcommand{\OpacityMLP}{\mathcal{A}}
\newcommand{\FeaturesMLP}{\mathcal{F}}
\newcommand{\Feature}{\mathbf{F}}
\newcommand{\ShaderMLP}{\mathcal{H}}
\newcommand{\Color}{\mathbf{C}}
\newcommand{\QuadraturePoints}{\mathcal{K}}
\newcommand{\Grid}{\mathcal{G}}
\newcommand{\StopGradient}{\cancel\nabla}
\newcommand{\Voxels}{\mathcal{B}}
\newcommand{\DualGrid}{\mathcal{G}}
\renewcommand{\triangle}{\mathbf{t}}

\newcommand{\pluggedin}{\includegraphics[scale=0.08]{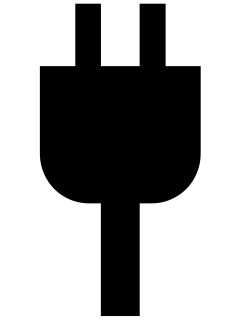}}

\newcommand{\ApproachName}{MobileNeRF}

\begin{document}

\title{\ApproachName{}: Exploiting the Polygon Rasterization Pipeline \\ for Efficient Neural Field Rendering on Mobile Architectures}
\author{
Zhiqin Chen$^{1,2,4}$ \quad
Thomas Funkhouser$^{1}$ \quad
Peter Hedman$^{1}$ \quad
Andrea Tagliasacchi$^{1,2,3,4}$ \\
\quad
\small{
Google Research\textsuperscript{1}
\quad
Simon Fraser University\textsuperscript{2}
\quad
University of Toronto\textsuperscript{3}
}
}
\maketitle
\footnotetext[4]{Work done while at Google.}
\begin{abstract}
Neural Radiance Fields (NeRFs) have demonstrated amazing ability to synthesize images of 3D scenes from novel views.   However, they rely upon specialized volumetric rendering algorithms based on ray marching that are mismatched to the capabilities of widely deployed graphics hardware.   This paper introduces a new NeRF representation based on textured polygons that can synthesize novel images efficiently with standard rendering pipelines.   The NeRF is represented as a set of polygons with textures representing binary opacities and feature vectors.  Traditional rendering of the polygons with a z-buffer yields an image with features at every pixel, which are interpreted by a small, view-dependent MLP running in a fragment shader to produce a final pixel color.   This approach enables NeRFs to be rendered with the traditional polygon rasterization pipeline, which provides massive pixel-level parallelism, achieving interactive frame rates on a wide range of compute platforms, including mobile phones.

Project page: \href{https://mobile-nerf.github.io}{https://mobile-nerf.github.io}

\end{abstract}

\section{Introduction}
\label{sec:intro}

Neural Radiance Fields (NeRF)~\cite{mildenhall2020nerf} have become a popular representation for novel view synthesis of 3D scenes.  They represent a scene using a multilayer perceptron (MLP) that evaluates a 5D implicit function estimating the density and radiance emanating from any position in any direction, which can be used in a volumetric rendering framework to produce novel images.   NeRF representations optimized to minimize multi-view color consistency losses for a set of posed photographs have demonstrated remarkable ability to reproduce fine image details for novel views.

One of the main impediments to wide-spread adoption of NeRF is that it requires specialized rendering algorithms that are poor match for commonly available hardware.  Traditional NeRF implementations use a volumetric rendering algorithm that evaluates a large MLP at hundreds of sample positions along the ray for each pixel in order to estimate and integrate density and radiance.  This rendering process is far too slow for interactive visualization.

\begin{figure}[t!]
\begin{center}
\includegraphics[width=1.0\linewidth]{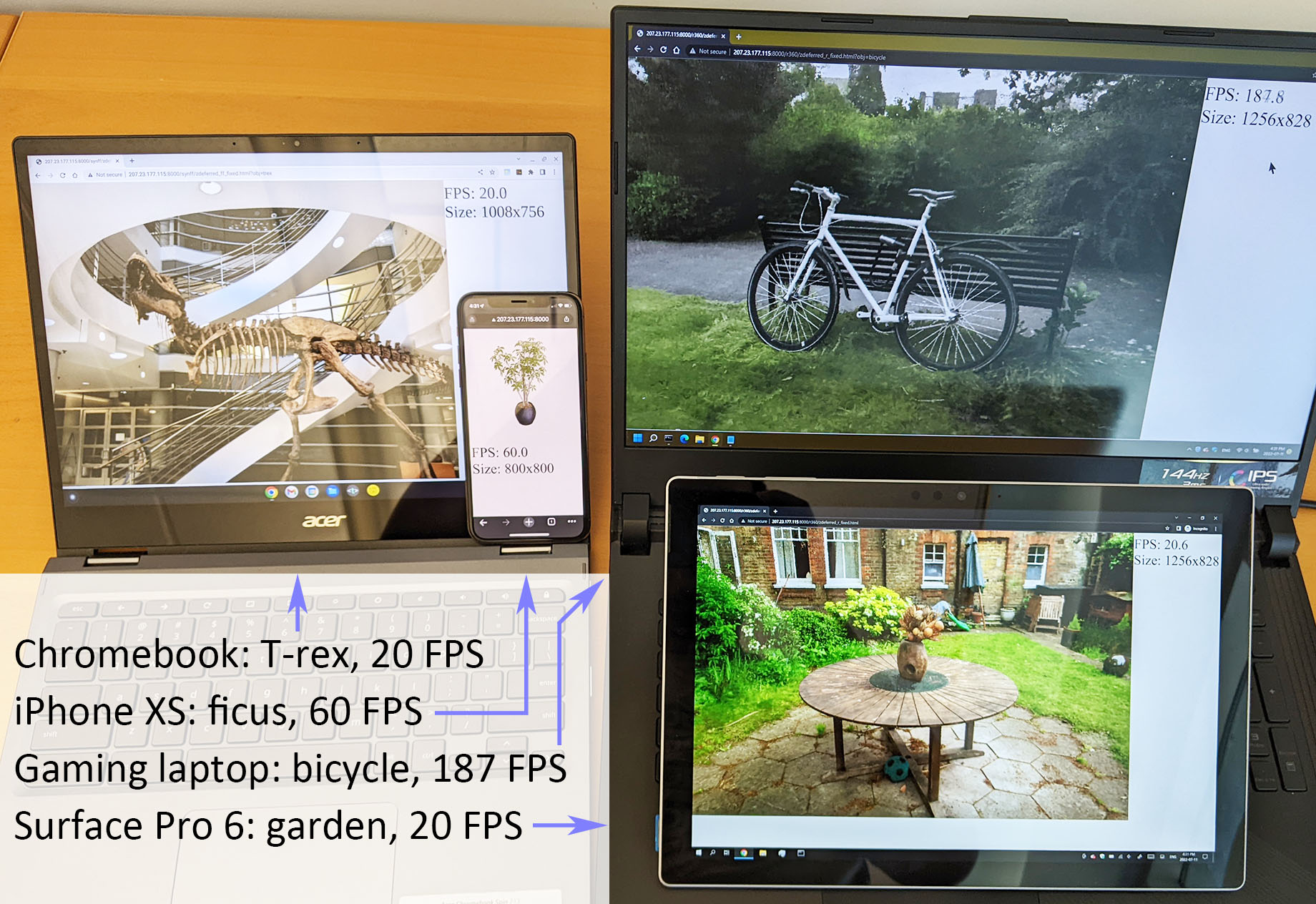}
\end{center}
\caption{
\textbf{Teaser --} 
We present a NeRF that can run on a variety of common devices at interactive frame rates.
}
\label{fig:teaser}
\end{figure} 

Recent work has addressed this issue by ``baking'' NeRFs into a sparse 3D voxel grid \cite{hedman2021snerg,Yu2021}.   For example, Hedman et al. introduced Sparse Neural Radiance Grids (SNeRG)~\cite{hedman2021snerg}, where each active voxel contains an opacity, diffuse color, and learned feature vector.  Rendering an image from SNeRG is split into two phases: the first uses ray marching to accumulate the precomputed diffuse colors and feature vectors along each ray, and the second uses a light-weight MLP operating on the accumulated feature vector to produce a view-dependent residual that is added to the accumulated diffuse color.  This precomputation and deferred rendering approach increase the rendering speed of NeRF by three orders of magnitude.   However, it still relies upon ray marching through a sparse voxel grid to produce the features for each pixel, and thus it cannot fully utilize the parallelism available in commodity graphics processing units (GPUs).
In addition, SNeRG requires a significant amount of GPU memory to store the volumetric textures, which prohibits it from running on common mobile devices.

In this paper, we introduce \ApproachName{}, a NeRF that can run on a variety of common mobile devices at interactive frame rates.
The NeRF is represented by a set of textured polygons, where the polygons roughly follow the surface of the scene, and the texture atlas stores opacity and feature vectors.  To render an image, we utilize the classic polygon rasterization pipeline with Z-buffering to produce a feature vector for each pixel and pass it to a lightweight MLP running in a GLSL fragment shader to produce the output color.
This rendering pipeline \textit{does not} sample rays or sort polygons in depth order, and thus can model only binary opacities.
However, it takes full advantage of the parallelism provided by z-buffers and fragment shaders in modern graphics hardware, and thus is $10\times$ faster than SNeRG with the same output quality on standard test scenes.
Moreover, it requires only a standard polygon rendering pipeline, which is implemented and accelerated on virtually every computing platform, and thus it runs on mobile phones and other devices previously unable to support NeRF visualization at interactive rates.

\paragraph{Contributions}
In summary, \ApproachName{}:
\begin{itemize}
\item Is $10\times$ \textit{faster} than the state-of-the-art~(SNeRG), with the same output quality;
\item Consumes less memory by storing \textit{surface} textures instead of volumetric textures, enabling our method to run on integrated GPUs with limited memory and power;
\item Runs on a web browser and is \textit{compatible} with all devices we have tested, as our viewer is an HTML webpage;
\item Allows real-time \textit{manipulation} of the reconstructed objects/scenes, as they are simple triangle meshes.
\end{itemize}
\section{Related work}
\label{sec:related}

Our work lies within the field of view-synthesis, which encompasses many areas of research: light fields, image-based rendering and neural rendering. To narrow the scope, we focus on methods that render output views in \textit{real-time}.

Light fields~\cite{Levoy96} and Lumigraphs~\cite{Gortler96} store a dense grid of images, enabling real-time rendering of high quality scenes, albeit with limited camera freedom and significant storage overhead.
Storage can be reduced by interpolating intermediate images with optical flow~\cite{Bertel2020}, representing the light field as a neural network~\cite{attal2022learning}, or by reconstructing a Multi-Plane Image~(MPI) representation of the scene~\cite{penner17,zhou18, mildenhall2019llff, flynn19, Wizadwongsa2021NeX}.
Multi-sphere images enable larger fields of view~\cite{broxton20, Attal2020}, but these representations still only support limited output camera motion 

Other approaches leverage explicit 3D geometry to enable more camera freedom. While early methods applied view-dependent texturing to a 3D mesh~\cite{debevec1998efficient, buehler2001unstructured, davis12}, later methods incorporated convolutional neural networks as a post-processing step to improve quality~\cite{MartinBrualla2018, hedman2018deep, thies2019neural}. Alternatively, the input geometry can be simplified into a collection of textured planes with alpha~\cite{lin2022neurmips}. Point-based representations further increase quality by jointly refining the scene geometry while training the post-processing network~\cite{lassner2021pulsar, ruckert2021adop, kopanas2021point}. 
However, as this convolutional post-processing runs independently per output frame it often results in a lack of 3D consistency.
Furthermore, unlike our work, they require powerful desktop GPUs and have not been demonstrated to run on a mobile device. Finally, unlike the vast majority of the methods above, our method does not need reconstructed 3D geometry as input.

It is also possible to extract explicit triangle meshes via differentiable inverse-rendering~\cite{gao2020deftet,munkberg2022extracting,Cole2021ICCV}.
DefTet~\cite{gao2020deftet} differentiably renders a tetrahedral grid with  occupancy and color at each vertex, and then compositing the interpolated values at all intersected faces along a ray.
NVDiffRec~\cite{munkberg2022extracting} combines differentiable marching tetrahedra~\cite{shen2021dmtet} with differentiable rasterization to perform full inverse rendering and extract triangle meshes, materials, and lighting from images.
This representation enables elaborate editing and scene relighting. However, it incurs a significant loss in view-synthesis quality. Furthermore, while real-time rendering is possible with simple lighting, global illumination (GI) is computationally infeasible on mobile hardware. In contrast, our method simply caches the outgoing radiance, which does not need expensive compute to model GI effects, and also results in higher view-synthesis quality.

\begin{figure*}[t!]
\begin{center}
\includegraphics[width=1.0\linewidth]{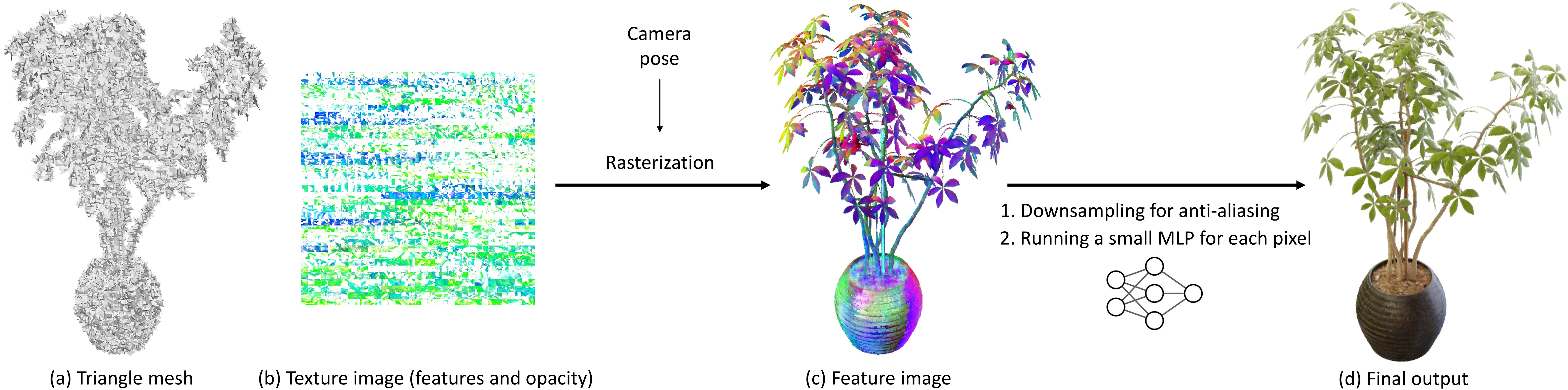}
\end{center}
\caption{
\textbf{Overview (rendering) --}
We represent the scene as a triangle mesh textured by deep features.
We first rasterize the mesh to a deferred rendering buffer.
For each visible fragment, we execute a neural deferred shader that converts the feature and view direction to the corresponding output pixel color.
}
\label{fig:overview_rendering}
\end{figure*}

NeRF~\cite{mildenhall2020nerf} represents the scene as a continuous field of opacity and view-dependent color, and produces images with volume rendering. This representation is 3D consistent and reaches high quality results~\cite{verbin2021refnerf,barron2022mipnerf360}.
However, rendering a NeRF involves evaluating a large neural network at multiple 3D locations per pixel, preventing real-time rendering.

Recent works have improved the training speed of NeRF.
For example, by modeling the opacity and color of entire ray segments instead of just points~\cite{lindell2021autoint} or by subdividing the scene and modeling each sub-region with a smaller neural network~\cite{rebain21}.
Recently, significant speed-ups have been achieved by decoding features fetched from a 3D embedding with a small neural network.
This embedding can either be a dense voxel grid~\cite{sun2021direct,Karnewar2022ReLUFields}, a sparse voxel grid~\cite{plenoxels}, a low-rank decomposition of a voxel grid~\cite{tensorf}, a point-based representation~\cite{xu2022point}, or a multi-resolution hash map~\cite{mueller2022instant}. 
These 3D embeddings can also be used without a trained decoder, for example by directly storing diffuse colors~\cite{lombardi19} or by encoding view-dependent colors as spherical harmonics~\cite{plenoxels}.
While these approaches drastically speed up training, they still require a large consumer GPU for rendering.

Rendering performance can further be increased by \textit{post-processing} a trained NeRF.
For example, by reducing the network queries per pixel with learned sampling~\cite{neff2021donerf}, by evaluating the network for larger ray segments~\cite{wu2022diver}, or by subdividing the scene into smaller networks~\cite{Reiser2021, wu2022snisr, rebain21}.
Alternatively, pre-computation can speed up rendering, by storing both scene opacity and a latent representation for view-dependent colors in a grid. 
FastNeRF~\cite{Garbin2021} uses a dense voxel grid and represents view-dependence with a global spherical basis function.
PlenOctrees~\cite{Yu2021} uses an octree representation, where each leaf node stores both opacity and spherical harmonics for colors.
SNeRG~\cite{hedman2021snerg} uses a sparse grid representation, and evaluates view-dependence as a post-process with a small neural network.
Among these real-time methods, only SNeRG has been shown to work on lower-powered devices without access to~CUDA. As our method directly targets rendering on low-powered hardware, we primarily compare with SNeRG in our experiments.

\begin{figure*}[t!]
\begin{center}
\includegraphics[width=1.0\linewidth]{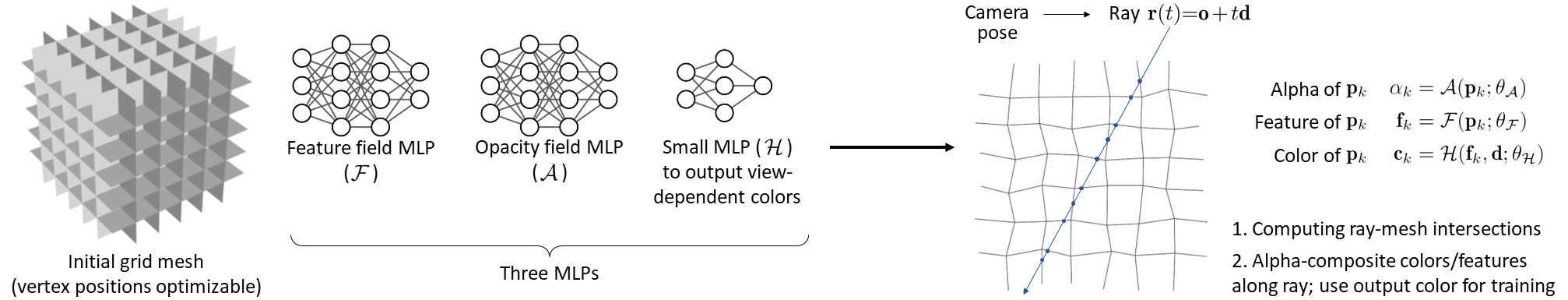}
\end{center}
\caption{
\textbf{Overview (train) -- } 
We initialize the mesh as a regular grid, and use MLPs to represent features and opacity for any point on the mesh.
For each ray, we compute its intersection points on the mesh, and alpha-composite the colors of those points to obtain the output color.
In a later training stage, we enforce \textit{binary} opacity, and perform super-sampling on features for anti-aliasing.
}
\label{fig:overview_train}
\end{figure*}

\section{Method}
\label{sec:method}
\label{sec:method_overview}
Given a collection of (calibrated) images, we seek to optimize a representation for \textit{efficient} novel-view synthesis.
Our representation consists of a polygonal mesh~(\Figure{overview_rendering}a) whose texture maps~(\Figure{overview_rendering}b) store features and opacity.
At rendering time, given a camera pose, we adopt a two-stage \textit{deferred rendering} process:
\begin{itemize}
\item \textbf{Rendering Stage 1} -- we rasterize the mesh to screen space and construct a \textit{feature image}~(\Figure{overview_rendering}c), i.e. we create a deferred rendering buffer in GPU memory;
\item \textbf{Rendering Stage 2} -- we convert these features into a color image via a (neural) deferred renderer running in a fragment shader, i.e. a small MLP, which receives a feature vector and view direction and outputs a pixel color~(\Figure{overview_rendering}d).
\end{itemize}
Our representation is built in three \textit{training} stages, gradually moving from a classical NeRF-like continuous representation towards a discrete one:
\begin{itemize}
\item \textbf{Training Stage 1 (\Section{stage1})} -- We train a NeRF-like model with \textit{continuous} opacity, where volume rendering quadrature points are derived from the polygonal mesh;
\item \textbf{Training Stage 2 (\Section{stage2})} -- We \textit{binarize} the opacities, as while classical rasterization can easily discard fragments, they cannot elegantly deal with semi-transparent fragments.
\item \textbf{Training Stage 3 (\Section{stage3})} -- We \textit{extract} a sparse polygonal mesh, bake opacities and features into texture maps, and store the weights of the neural deferred shader.
\end{itemize}
The mesh is stored as an OBJ file, the texture maps in PNGs, and the deferred shader weights in a (small) JSON file.
As we employ the standard GPU rasterization pipeline, our real-time renderer is simply an HTML webpage.

As representing continuous signals with discrete representations can introduce aliasing, we also detail a simple, yet computationally efficient, anti-aliasing solution based on super-sampling (\Section{antialiasing}).

\subsection{Continuous training (Training Stage 1)}
\label{sec:stage1}
As \Figure{overview_train} shows, our \textit{training setup} consists of a polygonal mesh $\Mesh{=}(\Triangles,\Vertices)$ and three MLPs.
The mesh topology $\Triangles$ is fixed, but the vertex locations $\Vertices$ and MLPs are optimized, similarly to NeRF, in an auto-decoding fashion by minimizing the mean squared error between predicted colors and ground truth colors of the pixels in the training images\floatersfootnote{}:
\begin{align}
\mathcal{L}_\mathbf{C} = \expect_{\ray}
\| \mathbf{C}(\ray) - \mathbf{C}_\text{gt}(\ray) \|_2^2.
\label{eq:rgb_reconstruction}
\end{align}
where the predicted color $\mathbf{C}(.)$ is obtained by alpha-compositing the radiance $\mathbf{c}_k$ along a ray $\mathbf{r}(t) {=} \mathbf{o} + t \mathbf{d}$, at the (depth sorted) quadrature points $\QuadraturePoints{=}\{t_k\}_{k=1}^K$:
\begin{align}
\mathbf{C}(\mathbf{r}) =
\sum_{k = 1}^{K}
T_k \alpha_k \mathbf{c}_k, \;\;\;\;
T_k = \prod_{l=1}^{k-1}(1-\alpha_l)
\label{eq:alpha_compositing}
\end{align}
where \textit{opacity} $\alpha_k$ and the view-dependent \textit{radiance} $\mathbf{c}_k$ are given by evaluating the MLPs at position $\point_k {=} \ray(t_k)$:
\begin{align}
\alpha_k &= \OpacityMLP(\mathbf{p}_k; \pars_\OpacityMLP)& \OpacityMLP &: \real^3 \rightarrow [0,1]
\label{eq:opacity}
\\
\mathbf{f}_k &= \FeaturesMLP(\mathbf{p}_k; \pars_\FeaturesMLP)& \FeaturesMLP &: \real^3 \rightarrow [0,1]^8
\label{eq:features}
\\
\mathbf{c}_k &= \ShaderMLP(\mathbf{f}_k,\mathbf{d}; \pars_\ShaderMLP)&  \ShaderMLP &: [0,1]^8 \times [-1,1]^3 \rightarrow [0,1]^3
\label{eq:color}
\end{align}
The small network $\ShaderMLP$ is our \textit{deferred neural shader}, which outputs the color of each fragment given the fragment feature and viewing direction.
Finally, note that \eq{alpha_compositing} does not perform compositing with volumetric density~\cite{mildenhall2020nerf}, but rather with opacity~\cite[Eq.8]{attal2022learning}.

\begin{figure}[t!]
\begin{center}
\includegraphics[width=1.0\linewidth]{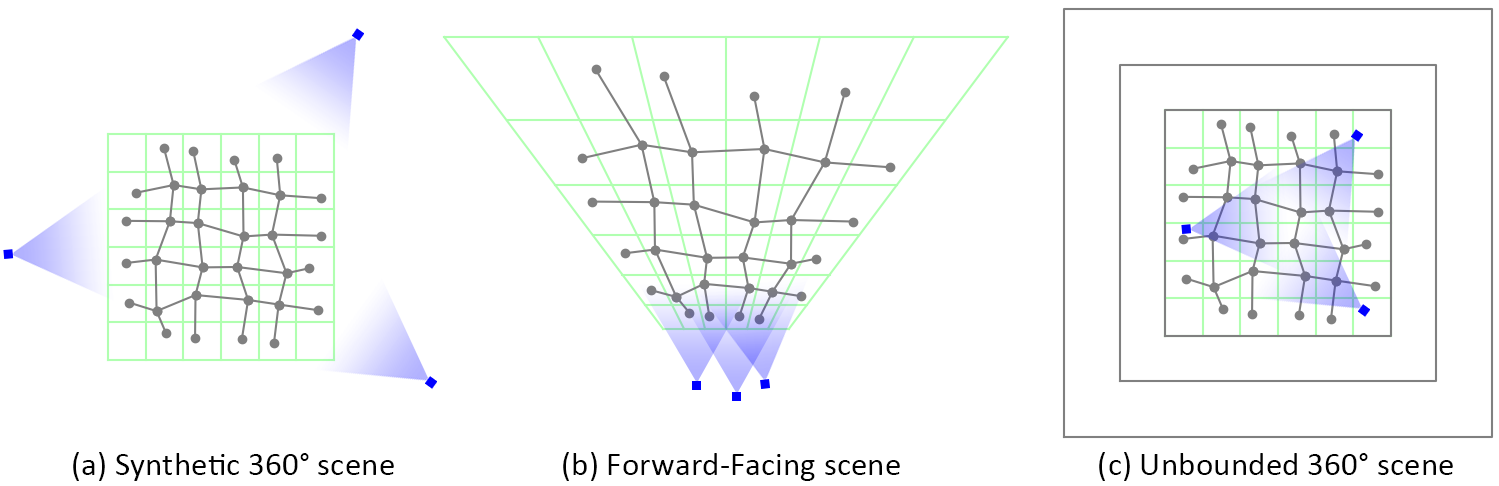}
\end{center}
\caption{
\textbf{Configurations of polygonal meshes --}
The meshes employed for the different types of scenes.
We sketch the distribution of camera poses in training views.
}
\label{fig:init_mesh}
\end{figure}

\paragraph{Polygonal mesh} 
Without loss of generality, we describe the polygonal mesh used in \textit{Synthetic $360^{\circ}$} scenes, and provide the configurations for \textit{Forward-Facing} and \textit{Unbounded $360^{\circ}$} scenes in \SupplementaryMaterial (\Section{supp_init_meshes}). 2D illustrations can be found in \Figure{init_mesh}.
We first define a \textit{regular} grid $\DualGrid$ of size $P {\times} P {\times} P$ in the unit cube centered at the origin; see \Figure{init_mesh}a.
We instantiate~$\Vertices$ by creating one vertex per voxel, and~$\Triangles$ by creating one quadrangle~(two triangles) per grid edge connecting the vertices of the four adjacent voxels, akin to Dual Contouring~\cite{ju2002dual,chen2022ndc}.
We locally parameterize vertex locations with respect to the voxel centers~(and sizes), resulting in $\Vertices {\in} [-.5, +.5]^{P {\times} P {\times} P {\times} 3}$ free variables.
During optimization, we initialize the vertex locations to $\Vertices{=}\mathbf{0}$, which corresponds to a regular~Euclidean lattice, and 
we regularize them to prevent vertices from exiting their voxels, and to promote their return to their neutral position whenever the optimization problem is under-constrained:
\begin{align}
\mathcal{L}_\Vertices = 
\sum_{\mathbf{v} \in \Vertices}
(10^{3}\,\mathcal{I}(\mathbf{v}) + 10^{-2}) \cdot ||\mathbf{v}||_1,
\end{align}
where the indicator function $\mathcal{I}(\mathbf{v}) {\equiv} 1$ whenever $\mathbf{v}$ is outside its corresponding voxel.

\begin{figure}[t]
\begin{center}
\includegraphics[width=1.0\linewidth]{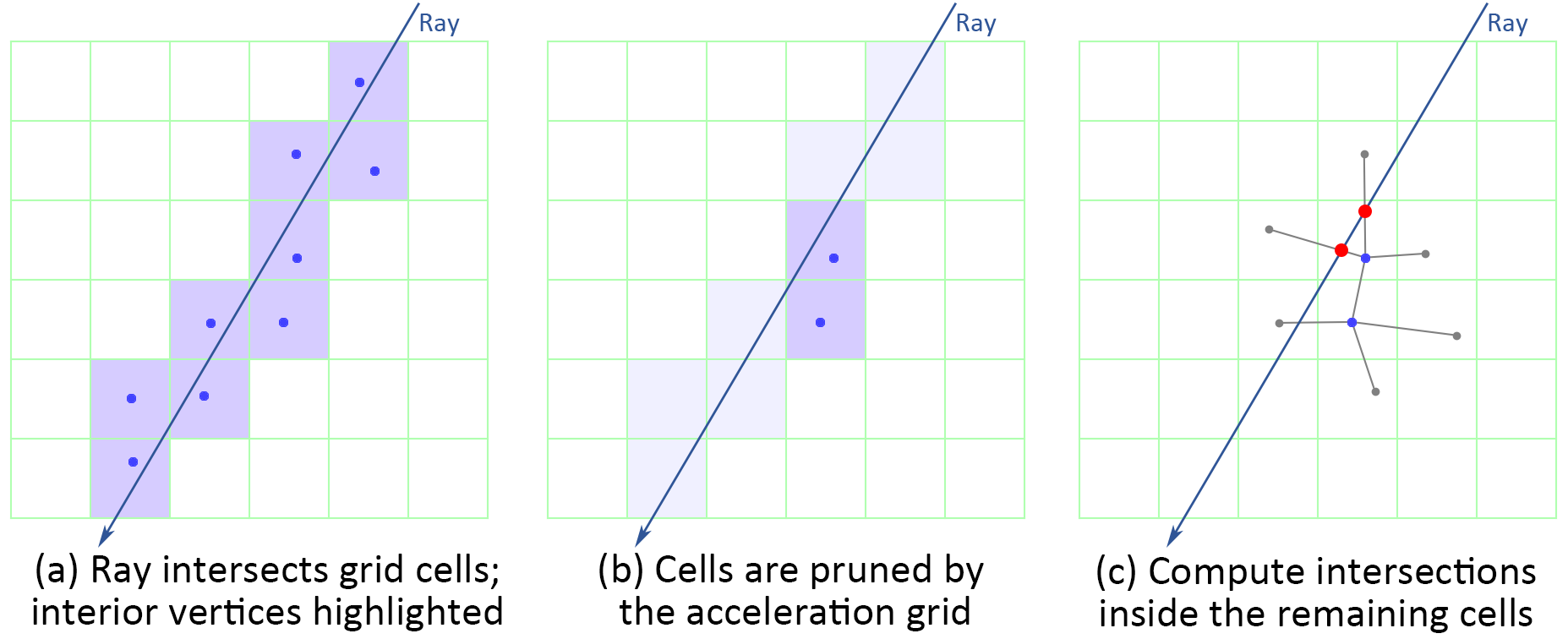}
\end{center}
\caption{
\textbf{Quadrature points --} are obtained by
(a) identifying cells that intersect the ray; 
(b) pruning cells that do not contain geometry; and,
(c) computing explicit intersections with the mesh.}
\label{fig:quadrature}
\end{figure}

\paragraph{Quadrature}
As evaluating the MLPs of our representation is computationally expensive, we rely on an acceleration grid to limit the cardinality $|\QuadraturePoints|$ of quadrature points.
First of all, quadrature points are only generated for the set of voxels that intersect the ray; see~\Figure{quadrature}a:
Then, like InstantNGP~\cite{mueller2022instant}, we employ an acceleration grid $\Grid$ to prune voxels that are unlikely to contain geometry; see~\Figure{quadrature}b.
Finally, we compute intersections between the ray and the faces of $\Mesh$ that are incident to the voxel's vertex to obtain the final set of quadrature points; see~\Figure{quadrature}c.
We use the barycentric interpolation to back-propagate the gradients from the intersection point to the three vertices in the intersected triangle.
For further technical details on the computation of intersections, we refer the reader to \SupplementaryMaterial (\Section{supp_quadrature}).
In summary, for each input ray $\ray$:
\begin{align}
\tilde\Voxels &= \text{intersect}(\ray, \DualGrid)
\label{eq:ray_voxel_intersection}
\\
\Voxels &= \{ b \in \tilde\Voxels ~|~ \Grid[b] > \tau_\Grid \}
\label{eq:acceleration}
\\
\QuadraturePoints &= \text{intersect}(\ray, \{ \triangle \in \Triangles ~|~ \triangle \cap \Voxels \})
\label{eq:ray_meshface_intersection}
\end{align}
where $(a\cap b){=}\text{\textit{true}}$ if $a$ intersects $b$, and the acceleration grid is supervised so to upper-bound\upperboundfootnote{} the alpha-compositing visibility $T_k \alpha_k$ \textit{across} viewpoints during training.
\begin{align}
\mathcal{L}_\Grid^\text{bnd} &= \sum_k \max( \StopGradient[T_k \alpha_k] - \Grid[\mathbf{p}_k] ,0)
\end{align}
where $\StopGradient[.]$ is the stop-gradient operator that prevents the acceleration grid from (negatively) affecting the image reconstruction quality.
This can be interpreted as a way to compute the so-called ``surface field'' \textit{during} NeRF training, as opposed to \textit{after} training as in nerf2nerf~\cite{nerf2nerf}.
Similarly to Plenoxels~\cite{plenoxels}, we additionally regularize the content of the grid by promoting its pointwise sparsity (i.e. lasso), and its spatial smoothness:
\begin{align}
\mathcal{L}_\Grid^\text{sparse} = \| \Grid \|_1^1
\quad \quad
\mathcal{L}_\Grid^\text{smooth} = \| \nabla \Grid \|_2^2
\end{align}

\subsection{Binarized training (Training Stage 2)}
\label{sec:stage2}
Rendering pipelines implemented in typical hardware \textit{do not} natively support semi-transparent meshes. Rendering semi-transparent meshes requires cumbersome (per-frame) sorting so to execute rendering in back-to-front order to guarantee correct alpha-compositing.
We overcome this issue by converting the smooth opacity~$\opacity_k{\in}[0,1]$ from \eq{opacity} to a discrete/categorical opacity $\dOpacity_k{\in}\{0,1\}$.
To optimize for discrete opacities via photometric supervision we employ a \textit{straight-through estimator}~\cite{straightthrough}:
\begin{align}
\dOpacity_k = \opacity_k + \StopGradient[ \mathbbm{1}(\alpha_k>0.5) - \alpha_k ]
\label{eq:dOpacity}
\end{align}
Please note that the gradients are transparently passed through the discretization operation (i.e.~$\nabla \dOpacity \equiv \nabla \opacity$), regardless of the values of $\alpha_k$ and the resulting $\dOpacity_k{\in}\{0,1\}$.
To stabilize training, we then co-train the continuous and discrete models:
\begin{align}
\mathcal{L}_\Color^\text{bin} &=
\expect_{\ray}
\| \hat\Color(\ray) - \Color_\text{gt}(\ray) \|_2^2 \\
\mathcal{L}_\Color^\text{stage2} &=
\tfrac{1}{2} \mathcal{L}_\Color^\text{bin}
~~+
\tfrac{1}{2} \mathcal{L}_\Color
\label{eq:MSE_plus_binary}
\end{align}
where $\hat\Color(\ray)$ is the output radiance corresponding to the discrete opacity model $\dOpacity$:
\begin{align}
\hat\Color(\ray) &= \sum_{k = 1}^{K}
\hat{T}_k \dOpacity_k \mathbf{c}_k, \;\;\;\;
\hat{T}_k = \prod_{l=1}^{k-1}(1-\dOpacity_l)
\label{eq:col_binarized}
\end{align}
Once \eq{MSE_plus_binary} has converged, we will apply a fine-tuning step to the weights in $\FeaturesMLP$ and $\ShaderMLP$ by minimizing $\mathcal{L}_\Color^\text{bin}$, while fixing the weights of others.


\subsection{Discretization (Training Stage 3)}
\label{sec:stage3}
After binarization and fine-tuning, we convert the representation into an explicit polygonal mesh (in OBJ format).
We only store quads if they are at least partially visible in the training camera poses (i.e. non-visible quads are discarded).
We then create a texture image whose size is proportional to the number of visible quads, and for each quad we allocate a $K{\times}K$ patch in the texture, similarly to Disney's Ptex~\cite{burley2008ptex}.
We use $K{=}17$ in our experiments, so that the quad has a $16{\times}16$ texture with half-a-pixel boundary padding.
We then iterate over the pixels of the texture, convert the pixel coordinate to 3D coordinates, and \textit{bake} the values of the discrete opacity~(i.e.~\eq{opacity} and \eq{dOpacity}) and features~(i.e.~\eq{features}) into the texture map.
We quantize the~$[0,1]$ ranges to 8-bit integers, and store the texture into (losslessly compressed) PNG images.
Our experiments show that quantizing the $[0,1]$ range with 8-bit precision, which is not accounted for during back-propagation, does not significantly affect rendering quality.

\subsection{Anti-aliasing}
\label{sec:antialiasing}
In classic rasterization pipelines, aliasing is an issue that ought to be considered to obtain high-quality rendering.
While classical NeRF hallucinates smooth edges via semi-transparent volumes, as previously discussed, semi-transparency would require per-frame polygon sorting. 
We overcome this issue by employing anti-aliasing by super-sampling.
While we could simply execute \eq{color} four times/pixel and average the resulting color, the execution of the deferred neural shader $\ShaderMLP$ is the computational bottleneck of our technique.
We can overcome this issue by simply averaging the features, that is, \textit{averaging the input} of the deferred neural shader, rather than averaging its output.
We first rasterize features (at $2\times$ resolution):
\begin{align}
\Feature(\ray) = \sum_k T_k \alpha_k \mathbf{f}_k,
\end{align}
and then average sub-pixel features to produce the anti-aliased representation we feed to our neural deferred shader:
\begin{align}
\Color(\ray) = \ShaderMLP
\left(
\expect_{\ray_\delta \sim \ray}[\Feature(\ray_\delta)], \:
\expect_{\ray_\delta \sim \ray}[\mathbf{d}_\delta]
\right)
\label{eq:antialiased_shader}
\end{align}
where $\expect_{\ray_\delta {\sim} \ray}$ computes the average between the sub-pixels~(i.e. four in our implementation), and $\mathbf{d}_\delta$ is the direction of ray $\ray_\delta$.
Note how with this change we only query~$\ShaderMLP$ \textit{once} per output pixel. Finally, this process is analogously applied to \eq{col_binarized} for discrete occupancies $\dOpacity$.
These changes for anti-aliasing are applied in training stage 2 \eq{MSE_plus_binary}.

\subsection{Rendering}
\label{sec:rendering}
The result of the optimization process is a textured polygonal mesh (where texture maps store features rather than colors) and a small MLP~(which converts view direction and features to colors).
Rendering this representation is done in two passes using a deferred rendering pipeline:
\begin{enumerate}
\item we rasterize all faces of the textured mesh with a z-buffer to produce a $2M {\times} 2N$ feature image with 12 channels per pixel, comprising $8$ channels of learned features, a binary opacity, and a 3D view direction;
\item we synthesize an $M \times N$ output RGB image by rendering a textured rectangle that uses the feature image as its texture, with linear filtering to average the features for antialiasing. We apply the small MLP for pixels with non-zero alphas to convert features into RGB colors. The small MLP is implemented as a GLSL fragment shader.
\end{enumerate}
These rendering steps are implemented within the classic rasterization pipeline.  Since z-buffering with binary transparency is order-independent, polygons \textit{do not} need to be sorted into depth-order for each new view, and thus can be loaded into a buffer in the GPU once at the start of execution.
Since the MLP for converting features to colors is very small, it can be implemented in a GLSL fragment shader~\cite{hedman2021snerg}, which is run in parallel for all pixels.
These classical rendering steps are highly-optimized on GPUs, and thus our rendering system can run at interactive frame rates on a wide variety of devices; see~\Table{test_FPS}.
It is also easy to implement, since it requires only standard polygon rendering with a fragment shader.
Our interactive viewer is an HTML webpage with Javascript, rendered by WebGL via the \texttt{threejs} library.

\begin{figure*}[t!]
\begin{center}
\includegraphics[width=1.0\linewidth]{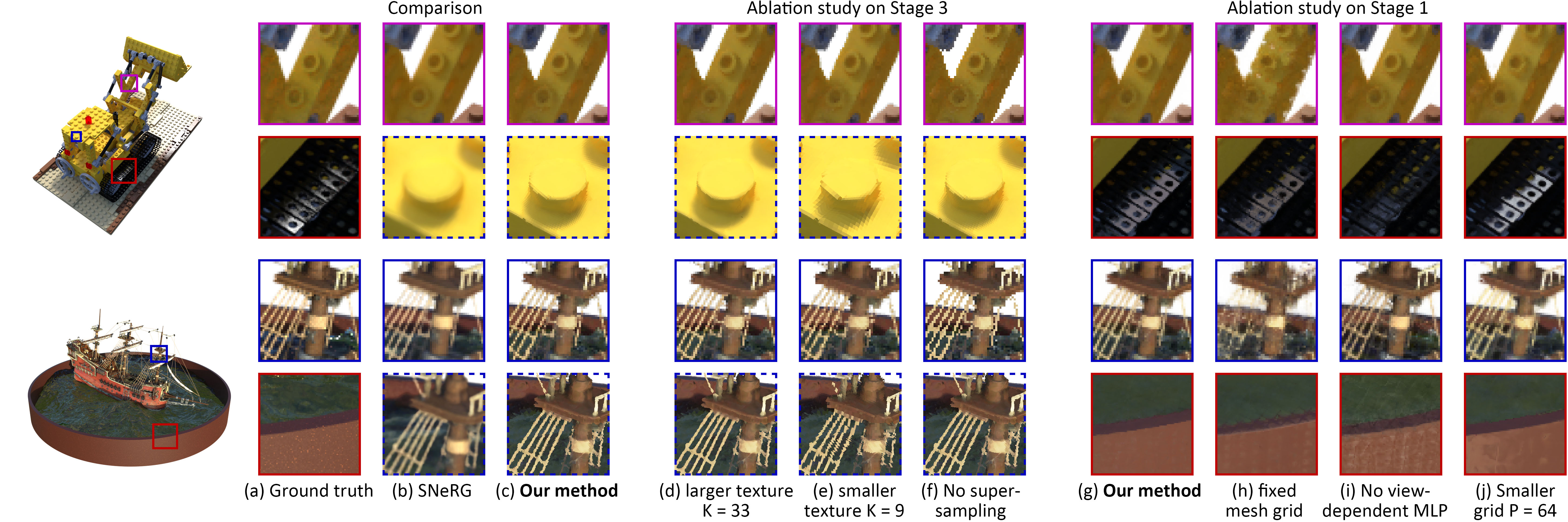}
\end{center}
\caption{
\textbf{Qualitative Results --}
Comparisons to the state-of-the-art and ablation studies. With a \textit{solid} line we denote zoom-ins of the rendered ($800 {\times} 800$) image, while with a \textit{dashed} line we move the camera to zoom-in onto the same detail.
}
\label{fig:qualitative}
\end{figure*} 

\section{Experiments}
\label{sec:exps}

We run a series of experiments to test how well \ApproachName{} performs on a wide variety of scenes and devices.   We test on three datasets: the 8 synthetic $360^{\circ}$ scenes from NeRF~\cite{mildenhall2020nerf}, the 8 forward-facing scenes from LLFF~\cite{mildenhall2019llff}, and 5 unbounded $360^{\circ}$ outdoor scenes from Mip-NeRF 360 ~\cite{barron2022mipnerf360}.  We compare with SNeRG~\cite{hedman2021snerg}, since, to our knowledge, it is the only NeRF model that can run in real-time on common devices. We also include extensive ablation studies to investigate the impact of different design choices.

\subsection{Comparisons}
To show the superior performance and compatibility of our method, we test our method and SNeRG on a variety of devices, as shown in~\Table{test_devices}. We report the rendering speed in~\Table{test_FPS}.
The rendering resolution is the same as the training images: $800{\times}800$ for synthetic, $1008{\times}756$ for forward-facing, and $1256{\times}828$ for unbounded.
We test all methods on a chrome browser and rotate/pan the camera in a full circle to render 360 frames.
Note that SNeRG is unable to represent unbounded $360^{\circ}$ scenes due to its regular grid representation, and it does not run on phone or tablet due to compatibility or out-of-memory issues.
We also report the GPU memory consumption and storage cost in \Table{test_space}.
\ApproachName{} requires 5x less GPU memory than SNeRG.

\begin{table}[!t]
\begin{center}
\resizebox{1.0\linewidth}{!}{
\begin{tabular}{lcccc}
\hline
Device & Type & OS & GPU & Power \\
\hline
iPhone XS & Phone & iOS 15 & Integrated GPU & 6W\\
Pixel 3 & Phone & Android 12 & Integrated GPU & 9W \\
Surface Pro 6 & Tablet & Windows 10 & Integrated GPU & 15W \\
Chromebook & Laptop & Chrome OS & Integrated GPU & 15W \\
Gaming laptop & Laptop & Windows 11 & NVIDIA RTX 2070 & 115W \\
Desktop & PC & Ubuntu 16.04 & NVIDIA RTX 2080 Ti & 250W \\
\hline
\end{tabular}}
\end{center}
\caption{\textbf{Hardware specs --} of the devices used in our rendering experiments.
The power is the max GPU power for discrete NVIDIA cards, and the combined max CPU and GPU power for integrated GPUs.}
\label{table:test_devices}
\end{table}
\begin{table}[!t]
\begin{center}
\resizebox{1.0\linewidth}{!}{
\begin{tabular}{l|cc|cc|c}
\hline
Dataset & \multicolumn{2}{c|}{Synthetic $360^{\circ}$} & \multicolumn{2}{c|}{Forward-facing} & \footnotesize{Unbounded $360^{\circ}$} \\
Method & Ours & SNeRG & Ours & SNeRG & Ours \\
\hline
iPhone XS & {\bf 55.89} & 0.0\footnotesize{$\frac{8}{8}$} & {\bf 27.19}\footnotesize{$\frac{2}{8}$} & 0.0\footnotesize{$\frac{8}{8}$} & 22.20\footnotesize{$\frac{4}{5}$} \\
Pixel 3 & {\bf 37.14} & 0.0\footnotesize{$\frac{8}{8}$} & {\bf 12.40} & 0.0\footnotesize{$\frac{8}{8}$} & 9.24   \\
Surface Pro 6 & {\bf 77.40} & \tiny{Unsupported} & {\bf 21.51} & \tiny{Unsupported} & 19.44 \\
Chromebook & {\bf 53.67} & 22.62\footnotesize{$\frac{2}{8}$} & {\bf 19.44} & 7.85\footnotesize{$\frac{3}{8}$} & 15.28 \\
Gaming laptop & {\bf 178.26} & 8.30\footnotesize{$\frac{1}{8}$} & {\bf 57.72} & 3.63 & 55.32 \\
Gaming laptop \pluggedin & {\bf 606.73} & 43.87\footnotesize{$\frac{1}{8}$} & {\bf 250.17} & 26.01 & 192.59  \\
Desktop \pluggedin & {\bf 744.91} & 207.26 & {\bf 349.34} & 50.71 & 279.70 \\
\hline
\end{tabular}
}
\end{center}
\caption{
\textbf{Rendering speed --}
on various devices in frames per second (FPS).
The devices are on battery, except for the gaming laptop and the desktop which are plugged in, indicated with a \pluggedin. The mobile devices (first four rows) have almost identical rendering speed when plugged in. With the notation $\frac{M}{N}$ we indicate that $M$ out of $N$ testing scenes failed to run due to out-of-memory errors.
}
\label{table:test_FPS}
\end{table}

\begin{table}[!t]
\begin{center}
\resizebox{1.0\linewidth}{!}{
\begin{tabular}{l|cc|cc|c}
\hline
Dataset & \multicolumn{2}{c|}{Synthetic $360^{\circ}$} & \multicolumn{2}{c|}{Forward-facing} & \footnotesize{Unbounded $360^{\circ}$} \\
Method & Ours & SNeRG & Ours & SNeRG & Ours \\
\hline
GPU memory & {\bf 538.38} & 2707.25 & {\bf 759.25} & 4312.13 & 1162.20 \\
Disk storage & 125.75 & {\bf 86.75} & {\bf 201.50} & 337.25 & 344.60 \\
\hline
\end{tabular}
}
\end{center}
\caption{
\textbf{Resources --} memory and disk storage~(MB).
}
\label{table:test_space}
\end{table}

\begin{table}[!t]
\begin{center}
\resizebox{1.0\linewidth}{!}{
\setlength{\tabcolsep}{2pt}
\begin{tabular}{l|ccc|ccc|ccc}
\hline
 & \multicolumn{3}{c|}{Synthetic $360^{\circ}$} & \multicolumn{3}{c|}{Forward-facing} & \multicolumn{3}{c}{Unbounded $360^{\circ}$} \\
 & PSNR$\uparrow$ & SSIM$\uparrow$ & LPIPS$\downarrow$ & PSNR$\uparrow$ & SSIM$\uparrow$ & LPIPS$\downarrow$ & PSNR$\uparrow$ & SSIM$\uparrow$ & LPIPS$\downarrow$ \\
\hline
NeRF & 31.00 & 0.947 & 0.081 & 26.50 & 0.811 & 0.250 & - & - & - \\
JAXNeRF & 31.65 & 0.952 & 0.051 & 26.92 & 0.831 & 0.173 & 21.46 & 0.458 & 0.515 \\

NeRF++ & - & - & - & - & - & - & 22.76 & 0.548 & 0.427 \\
\hline
SNeRG & 30.38 & {\bf 0.950} & {\bf 0.050} & 25.63 & 0.818 & {\bf 0.183} & - & - & - \\
Ours & {\bf 30.90} & 0.947 & 0.062 & {\bf 25.91} & {\bf 0.825} & {\bf 0.183} & 21.95 & 0.470 & 0.470  \\
\hline
\end{tabular}
}
\end{center}
\caption{
\textbf{Quantitative Analysis --}
For NeRF~\cite{mildenhall2020nerf} and NeRF++~\cite{nerf++}, we dash entries where the original papers did not report quantitative performance.
For SNeRG, while one could extend the method to include the unbounded design from~\cite{barron2022mipnerf360}, implementing this is far from trivial.
Our method can be easily adapted to work across all modalities.
}
\label{table:test_image}
\end{table}

\paragraph{Rendering quality}
We report the rendering quality in \Table{test_image}, while comparing with other methods using the common PSNR, SSIM~\cite{SSIM}, and LPIPS~\cite{LPIPS} metrics.
Our method has roughly the same image quality as SNeRG, and better than NeRF.
Visual results are shown in \Figure{qualitative} (a-c).
Our method achieves image quality similar to SNeRG when the camera is at an appropriate distance. When the camera is zoomed in, SNeRG tends to render over-smoothed images.

\begin{table}[!t]
\begin{center}
\resizebox{1.0\linewidth}{!}{
\begin{tabular}{l|cc|cc|cc}
\hline
 & \multicolumn{2}{c|}{Synthetic $360^{\circ}$} & \multicolumn{2}{c|}{Forward-facing} & \multicolumn{2}{c}{Unbounded $360^{\circ}$} \\
 & V & T & V & T & V & T \\
\hline
Number & 494,289 & 224,341 & 830,076 & 338,535 & 1,436,033 & 608,785 \\
Percentage & 1.964\% & 1.783\% & 3.298\% & 2.690\% & 4.891\% & 4.147\% \\
\hline
\end{tabular}
}
\end{center}
\caption{
\textbf{Polygon count --}
Average number of vertices and triangles produced, and their percentage compared to all available vertices/triangles in the initial mesh.
}
\label{table:test_v_t_count}
\end{table}

\paragraph{Polygon count}
\Table{test_v_t_count} shows the average number of vertices and triangles produced by our method, and the percentage compared to all available vertices/triangles in the initial mesh. As we only retain visible triangles, most vertices/triangles are removed in the final mesh.

\begin{figure}[t!]
\begin{center}
\includegraphics[width=1.0\linewidth]{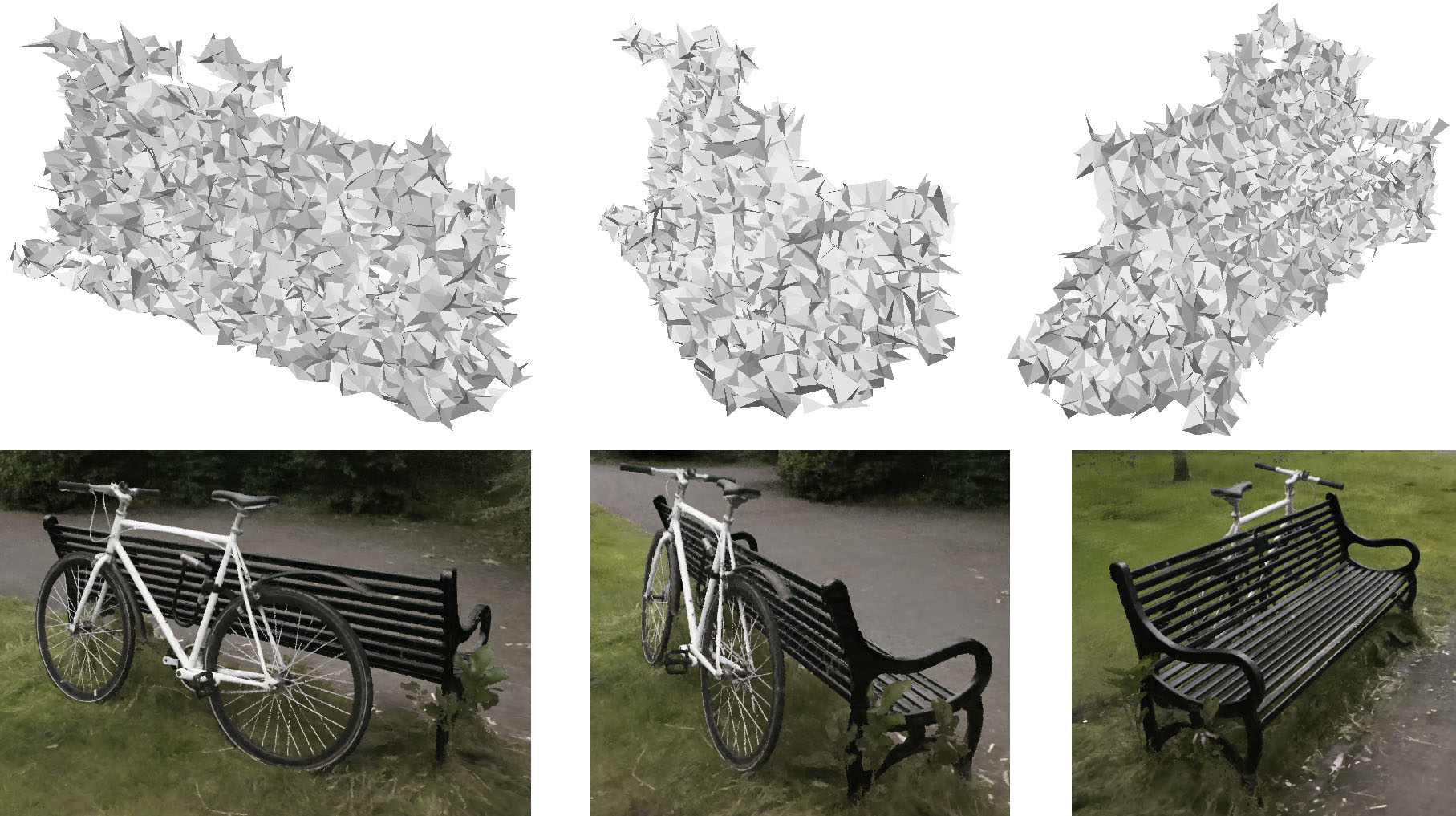}
\end{center}
\caption{
\textbf{Shading mesh} -- not textured.
The mesh corresponds to the bicycle (see \Figure{teaser}). We manually removed the background mesh to better show the geometry of the object. Zoom-in to see more details. In the bottom, we also show the rendered images of our method. Note how the coarse mesh is able to represent detailed structures such as the spokes of the wheels and the wires around the handles, thanks to high-resolution textures with transparencies.
}
\label{fig:mesh_noshade}
\end{figure}

\paragraph{Shading mesh}
In \Figure{overview_rendering}a and \Figure{mesh_noshade}, we show the extracted triangle meshes without the textures. Most triangle faces do not align with the actual object surface. This is perhaps due to the ambiguity that good rendering quality can be achieved despite how the triangles are aligned.
For example, the results of our method after Stage~1 in \Table{ablation_image} is similar to other methods in \Table{test_image}.
Therefore, better regularization losses or training objectives need to be devised if one wishes to have better surface quality. However, optimizing vertices does improve the rendering quality, as shown in~\Figure{qualitative}h.

\begin{table}[!t]
\begin{center}
\resizebox{1.0\linewidth}{!}{
\begin{tabular}{l|cc|cc}
\hline
 & \multicolumn{2}{c|}{Synthetic $360^{\circ}$} & \multicolumn{2}{c}{Forward-facing} \\
 & PSNR$\uparrow$ & SSIM$\uparrow$ & PSNR$\uparrow$ & SSIM$\uparrow$ \\
\hline
Stage 1, {\bf our method} & {\bf 32.13} & {\bf 0.955} & 26.57 & {\bf 0.839} \\
Stage 1, fixed mesh grid & 29.87 & 0.938 & 25.43 & 0.797 \\
Stage 1, no view-dependent MLP & 29.91 & 0.935 & 25.91 & 0.824 \\
Stage 1, smaller grid $P{=}128 \rightarrow 64$ & 31.58 & 0.952 & 26.39 & 0.831 \\
Stage 1, no acceleration grid & 31.77 & 0.953 & {\bf 26.61} & 0.835 \\
\hline
Stage 2, {\bf our method} & {\bf 31.01} & {\bf 0.948} & {\bf 26.32} & {\bf 0.833} \\
Stage 2, no fine-tuning & 30.80 & 0.946 & 26.25 & 0.832 \\
Stage 2, only pseudo-gradients & 29.70 & 0.935 & 26.01 & 0.820 \\
Stage 2, binary loss & 30.89 & 0.947 & {\bf 26.32} & 0.832 \\
\hline
Stage 3, {\bf our method} & 30.90 & 0.947 & 25.91 & 0.825 \\
Stage 3, larger texture $K{=}17 \rightarrow 33$ & {\bf 30.99} & {\bf 0.948} & {\bf 26.14} & {\bf 0.830} \\
Stage 3, smaller texture $K{=}17 \rightarrow 9$ & 30.49 & 0.945 & 24.85 & 0.796 \\
Stage 3, no supersampling & 29.26 & 0.937 & 24.88 & 0.799 \\
\hline
\end{tabular}
}
\end{center}
\caption{
\textbf{Ablation -- } rendering quality. 
}
\label{table:ablation_image}
\end{table}

\begin{table}[!t]
\begin{center}
\resizebox{1.0\linewidth}{!}{
\begin{tabular}{l|ccc|cc}
\hline
 & \multicolumn{3}{c|}{Speed in FPS} & \multicolumn{2}{c}{Space in MB} \\
\hline
Synthetic $360^{\circ}$ scenes
 & Pixel 3 & Surface & Gaming & GPU    & Disk  \\
 &         & Pro 6   & laptop \pluggedin & memory & storage  \\
\hline
{\bf our method} & 37.14 & 77.40 & 606.73 & 538.38 & 125.75 \\
Larger texture $K=33$ & 32.48\footnotesize{$\frac{2}{8}$} & 59.15 & 589.20 & 1290.88 & 283.50 \\
Smaller texture $K=9$ & 37.74 & 94.62 & 617.74 & {\bf 336.63} & {\bf 67.00} \\
No supersampling & 51.81 & {\bf 113.41} & {\bf 649.86} & 440.25 & 125.75 \\
No view-dependent MLP & {\bf 52.16} & 96.76 & 638.30 & 538.38 & 125.75 \\
\hline
 & & & & & \\
\hline
Forward-facing scenes
 & Pixel 3 & Surface & Gaming & GPU    & Disk  \\
 &         & Pro 6   & laptop \pluggedin & memory & storage  \\
\hline
{\bf our method} & 12.40 & 21.51 & 250.17 & 759.25 & 201.50  \\
Larger texture $K=33$ & 12.88\footnotesize{$\frac{3}{8}$} & 18.79 & 241.52 & 2024.13 & 462.75 \\
Smaller texture $K=9$ & 12.70 & 23.61 & 257.64 & {\bf 394.13} & {\bf 105.75} \\
No supersampling & 16.97 & {\bf 42.11} & {\bf 413.02} & 645.00 & 201.50 \\
No view-dependent MLP & {\bf 23.72} & 28.06 & 385.65 & 759.25 & 201.50 \\
\hline
\end{tabular}
}
\end{center}
\caption{
\textbf{Ablation -- } rendering speed/memory.
}
\label{table:ablation_FPS_space}
\end{table}

\begin{figure}[t!]
\begin{center}
\includegraphics[width=1.0\linewidth]{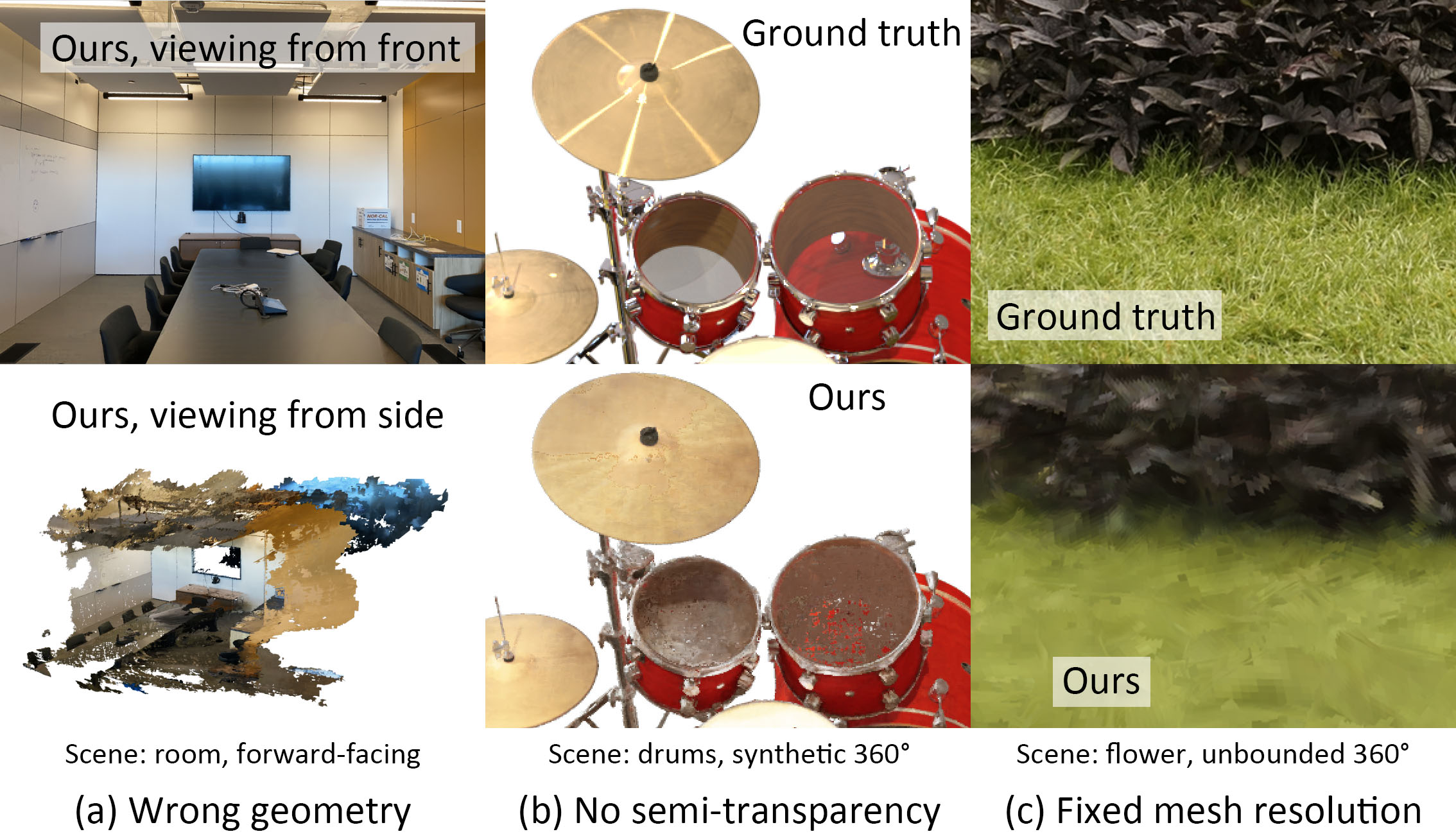}
\end{center}
\caption{
\textbf{Limitations --}
(a) the monitor/table are hollow, because the reflections are modelled as real objects behind the monitor and below the table. (b) our method generates scattered small fragments in the semi-transparent parts. (c) the camera is too close to the scene and details in the grass cannot be represented at the chosen texture resolution.
}
\label{fig:limitation}
\end{figure} 
\begin{figure}[t!]
\begin{center}
\includegraphics[width=1.0\linewidth]{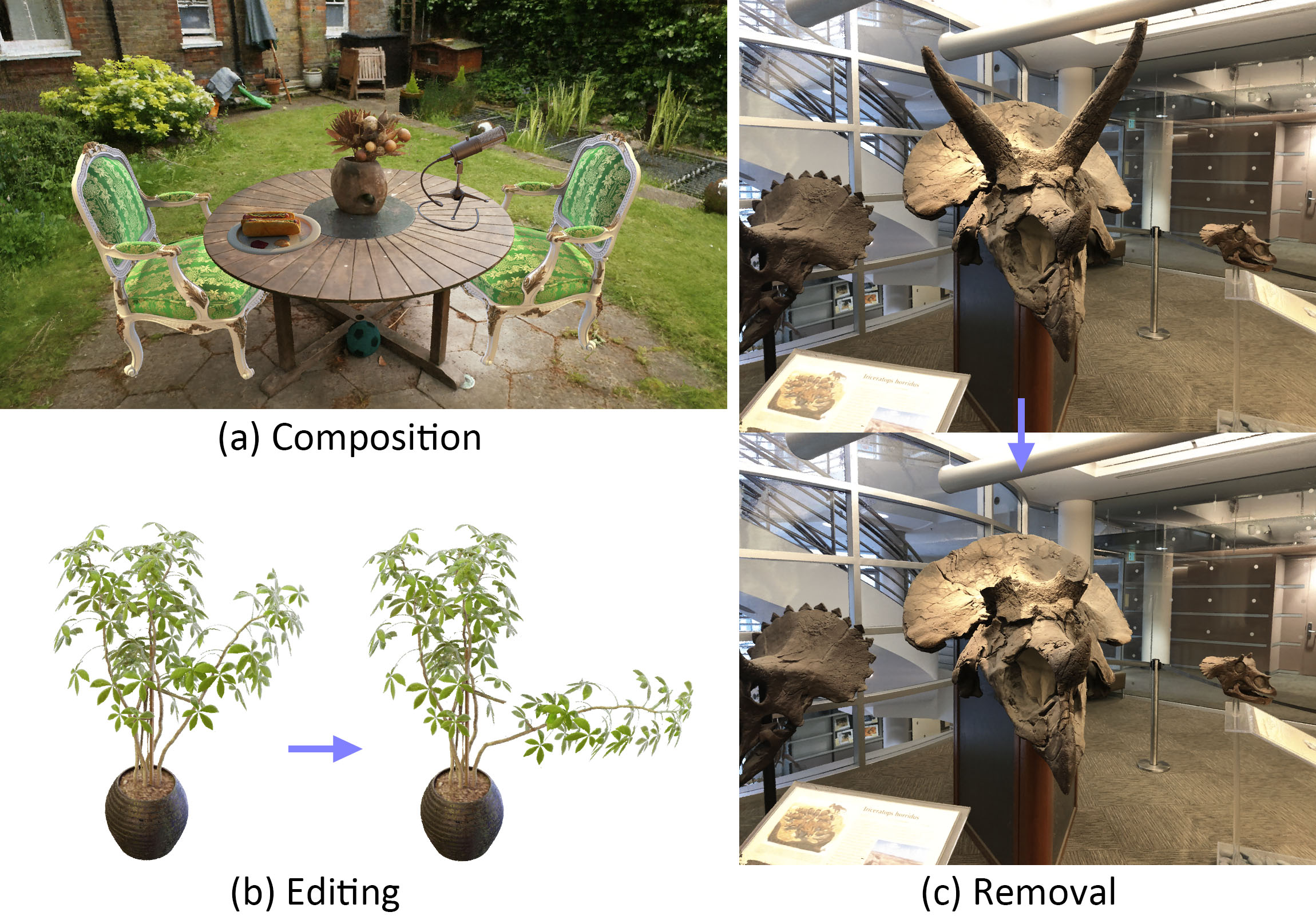}
\end{center}
\caption{
\textbf{Scene editing --}
(a) four objects learned from the synthetic scenes are added into an unbounded scene.
(b) a branch of the ficus is bent.
(c) the horns are removed.}
\label{fig:scene_editing}
\end{figure} 

\subsection{Ablation studies}
In \Table{ablation_image}, we show the rendering quality of our method at each stage, and report our ablation studies.
The rendering quality gradually drops after each stage, because each stage adds more constraints to the model.
In Stage~1, the performance drops significantly if we use a fixed regular grid mesh instead of having optimizable mesh vertices, or if we forgo view-dependent effects by directly predicting the color and alpha of each point.
The performance drops slightly if the grid is smaller~($P{=}64$ vs.~$128$).
If we remove the acceleration grid, we are not able to quadruple the batch size during training; the performance drops if we train this model the same number of iterations as our method.
Note that the PSNR of this model is higher on forward-facing scenes.
This is because the acceleration grid will remove cells that are not visible in the training images, thus cannot ``inpaint'' the objects and may leave holes.
In Stage 2, if we do not perform the fine-tuning step that only optimizes $\mathcal{F}$ and $\mathcal{H}$ and fix the weights of others, the performance drops.
If we only use the binary opacity with pseudo-gradients by applying $\mathcal{L}_\Color^\text{stage2}{=}\mathcal{L}_\Color^\text{bin}$ instead of Eq.~\ref{eq:MSE_plus_binary}, the performance drops.
If we use a binary loss on the predicted opacity, e.g., $\mathcal{L}_{binary}{=}-\sum |\alpha_k - 0.5| $, instead of using the pseudo-gradients with $\hat\Color(\ray)$, the performance drops slightly.
In stage 3, when we use a larger texture size $K{=}33$ instead of $17$, the performance improves, but the texture size will be quadrupled; the performance drops when we use a smaller texture size $K{=}9$.
If we remove the super-sampling step, the performance drops significantly.
Visual results are shown in \Figure{qualitative}.
We omit some models because they do not have significant visual differences compared to our method.
Notice the squared pixels of the texture images are clearly visible in the dashed-line boxes in (e) and almost invisible in (d).
The aliasing artifacts are conspicuous in the solid-line boxes in (f).
In Stage 1, if the grid vertices cannot be optimized, the results will be significantly worse, as shown in (h).
Without the small MLP, the model cannot handle reflections, as shown in (i).
Changing to a smaller grid size introduces some minor artifacts in (j).
In \Table{ablation_FPS_space}. we show the rendering speed and space cost if we use a larger or smaller texture size, or if we remove the super-sampling step, or if we only perform the rasterization without using the small MLP to predict the view-dependent colors. One can find that the super-sampling step and the small MLP have the most significant impact.

\section{Conclusions}
\label{sec:conclusions}
We introduce \ApproachName{}, an architecture that takes advantage of the classical rasterization pipeline (i.e. z-buffers and fragment shaders) to perform efficient rendering of surface-based neural fields on a wide range of compute platforms.
It achieves frame rates an order of magnitude faster than the previous state-of-the-art~(SNeRG) while producing images of equivalent quality.

\paragraph{Limitations}
Our estimated \textit{surface} may be incorrect, especially for scenes with specular surfaces and/or sparse views (\Figure{limitation}a);
it uses \textit{binary} opacities to avoid sorting polygons, and thus cannot handle scenes with semi-transparencies~(\Figure{limitation}b);
it uses fixed mesh and texture resolutions, which may be too coarse for close-up novel-view synthesis~(\Figure{limitation}c);
it models a radiance field without explicitly decomposing illumination and reflectance, and thus does not handle glossy surfaces as well as recent methods~\cite{verbin2021refnerf}.
Extending the polygon rendering pipeline with efficient partial sorting, levels-of-detail, mipmaps, and surface shading should address some of these issues.
Also, the current training speed of \ApproachName{} is slow due to NeRF’s MLP backbone. The extension of \ApproachName{} to fast-training architectures (e.g., Instant NGP~\cite{mueller2022instant}) constitutes an exciting avenue for future works.

The explicit mesh representation provided by \ApproachName{} gives us direct editing control over the NeRF model without any complex architectural change (e.g. ControlNerf~\cite{lazova2022control}), but in this paper we only superficially investigated these possibilities; see~\Figure{scene_editing} and the videos in the \SupplementaryMaterial (\Section{supp_scene_edit}).

\paragraph{Acknowledgements}
We thank the reviewers as well as Simon Kornblinth, Ting Chen, Daniel Rebain, Kevin Swersky, and David Fleet for their valuable feedback.
\clearpage

{\small
\bibliographystyle{ieee_fullname}
\bibliography{egbib}
}

\twocolumn[
\centering
\textbf{\Large\ApproachName{}: Exploiting the Polygon Rasterization Pipeline \\ for Efficient Neural Field Rendering on Mobile Architectures} \\
\vspace{0.5em}\Large{(Supplementary Material)} \\
\vspace{1.0em}
] 
\setcounter{page}{1}

\appendix

\section{More results}

\begin{figure*}[t!]
\begin{center}
\includegraphics[width=0.6\linewidth]{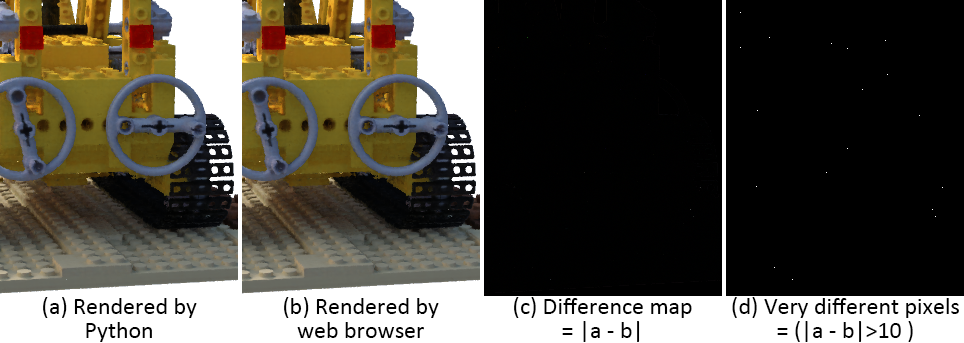}
\end{center}
\caption{
\footnotesize{Comparison between images rendered in Python and in a web browser. Image pixel value range is 0-255. Zoom in for details.}
}
\label{fig:python_vs_browser}
\end{figure*} 

Check out our project page: \href{https://mobile-nerf.github.io}{https://mobile-nerf.github.io}

The models used in the online demos of our project page are the same as the ones used in our paper. The rendered images of our method are nearly identical whether they are rendered in Python (for computing quantitative metrics) or web browsers, see Figure \ref{fig:python_vs_browser}. If one overlays the difference image and the rendered image, one can find that the few very different pixels are all on the boundary of a part, which indicates that they are likely caused by precision errors in rasterization.

\section{[Post-submission] Shader code optimization}
\label{sec:supp_new_stuff}

With the optimizations suggested by Noeri Huisman, we have greatly improved the rendering speed of the fragment shader containing the small MLP $\mathcal{H}$. specifically, we
\begin{itemize}
\item Inject network weights directly into the shader source code, instead of using weight textures and texel fetches;
\item Use mat4 and vec3 multiplications in all operations, instead of multiplying and adding float numbers.
\end{itemize}
When tested on the 5 real unbounded scenes with a Samsung Galaxy S22 Ultra mobile phone, the average FPS is 35 using the optimized implementation, which is 35\% faster than our original implementation (26 FPS).

Note that our default rendering setting is \textbf{deferred rendering}, that is, meshes and textures are rasterized into a feature image with pixels containing features, and then we treat this feature image as the texture image of a rectangle polygon mesh and rasterize it into the final output image with pixels containing RGB colors using our MLP fragment shader.
This design was to ensure that the MLP shader is executed once per output pixel.
in contrast, \textbf{forward rendering}, where the meshes and textures are directly rasterized into pixels containing RGB colors using the MLP fragment shader, will need to execute the MLP shader once per fragment. Since the number of fragments are usually much larger than the number of output pixels, the deferred rendering setting has a speed advantage over forward rendering when the speed of the MLP shader is a bottleneck in rendering.

Now that the optimized MLP fragment shader is so fast, we have observed that forward rendering is much faster than deferred rendering on some mobile devices.
When tested on the 5 real unbounded scenes with a Samsung Galaxy S22 Ultra mobile phone, the average FPS is 84 using the optimized forward rendering implementation, which is 223\% faster than our original deferred rendering implementation (26 FPS) and 140\% faster than the optimized deferred rendering implementation (35 FPS).

Note that forward rendering is still slower than deferred rendering on some devices due to excessive overdraw. A depth pre-pass can help resolve this issue, but it may not be available on certain devices.

We provide demos for both forward and deferred rendering settings on our project page: \href{https://mobile-nerf.github.io}{https://mobile-nerf.github.io}.
However, the models in these demos were trained with deferred rendering, where super-sampling is done on pixel features, while the super-sampling in forward rendering should be done on pixel colors. Therefore the rendered results may be slightly different.

\section{Scene editing}
\label{sec:supp_scene_edit}

Our representation is a textured mesh with baked lighting, and thus can be used in any application that combines, renders, or manipulates such meshes.  \Figure{scene_editing} (a) shows a simple example where meshes learned from four different sets of photos are composited into a single scene. The scene, rendered in $1920{\times}1080$ resolution without super-sampling, runs at 150 FPS on the gaming laptop, and consumes 1.5 GB of GPU memory.
Similarly \Figure{scene_editing} (b)(c) show scenes where some parts or objects are edited or removed by manipulating the triangle meshes of the scenes in a 3D modeling software. 
The resulting renders do not account for differences in illumination between the captured photos or indirect illumination between different meshes.  However, it suggests an easy way to create ``photorealistic-looking'' scenes from a library of objects captured using photos rather than painstaking 3D modeling. 

\href{https://youtu.be/kVy2W6afuyk}{https://youtu.be/kVy2W6afuyk} shows three examples where we manipulate the learned NeRF objects interactively in real-time. We also highlight how easy it is to implement these operations with our mesh representation. In contrast, implementing those with classic NeRF is non-trivial.

In the first example, we render all 8 objects learned from the synthetic scenes at the same time, and we move the objects by using mouse to drag the objects. This is implemented by a single line of code with the \emph{DragControls} class provided in the \emph{threejs} library. \emph{DragControls} is designed for manipulating meshes, which suits our needs exactly since our objects are meshes. We also cast real-time shadow of the objects by applying shadow mapping. This is implemented by having a directional light, an ambient light, and a plane below the objects to receive shadows. The drag control and the real-time shadow are also used in the following examples.

In the second example, we interactively deform the learned chair object to create new variations of chairs. To implement the deformation, we only need to deform the vertex positions of the meshes, and this is achieved by adding vertex deformation code in the vertex shader. Specifically, we implemented three operations: moving the chair up/down will lengthen or shorten its legs, moving the chair left/right will adjust its width, and moving the chair forward/backward will adjust the skew of its back.

In the third example, we render 9 ficus objects, which are considered  ``NeRF'' objects, and a blue ball, which is a classic object with standard material used in classic rendering. We again change the vertex shader to make the leaves of the plants to be repelled by the blue ball.

\section{Training}
Our training stages are formalized as follows.
In the first training stage, we optimize
\begin{align}
\argmin_{\Vertices, \pars_\OpacityMLP, \pars_\FeaturesMLP, \pars_\ShaderMLP} \quad \mathcal{L}_\Color + w_d \mathcal{L}_\text{dist} + \mathcal{L}_\text{v}
\end{align}
and
\begin{align}
\argmin_{\Grid} \quad \mathcal{L}_\Grid^\text{bnd} + w_{g1} \mathcal{L}_\Grid^\text{sparse} + w_{g2} \mathcal{L}_\Grid^\text{smooth},
\label{eq:acc_grid_loss}
\end{align}
where $w_{g1} = w_{g2} = 10^{-5}$. $w_d$ is set to $0.0$ for synthetic $360^{\circ}$ scenes, $0.01$ for forward-facing scenes, and $0.001$ for unbounded $360^{\circ}$ scenes.
In the second training stage, we optimize
\begin{align}
\argmin_{\Vertices, \pars_\OpacityMLP, \pars_\FeaturesMLP, \pars_\ShaderMLP} \quad \mathcal{L}_\Color^\text{stage2} + w_d \mathcal{L}_\text{dist} + \mathcal{L}_\text{v}
\end{align}
and Eq.~\ref{eq:acc_grid_loss}. When the loss converges, we fix the weights of $\Vertices$, $\pars_\OpacityMLP$, and $\Grid$ and optimize
\begin{align}
\argmin_{\pars_\FeaturesMLP, \pars_\ShaderMLP} \quad \mathcal{L}_\Color^\text{bin}.
\end{align}

\section{Network architectures}
We adopt the MLP designed in NeRF as the network for both $\mathcal{A}$ and $\mathcal{F}$. We increase the hidden layer sizes from $256$ to $384$, since $\mathcal{A}$ and $\mathcal{F}$ are not used during inference, so we can afford more time on training.
The small MLP $\mathcal{H}$ is the same as the small MLP used in SNeRG, with two hidden layers, each consisting 16 neurons.

\section{More details about texture images}
Since the features to be stored are 8-dimensional, we use two PNG images to store them. Each PNG image has 4 channels, therefore two PNG images have a total of 8 channels to store 8-d features.
To avoid having an extra image to store the binary alpha (opacity) channel, we squeeze the alpha channel into the first feature channel, so that the alpha is one when the first feature channel is non-zero, and zero when the channel is zero.
Since phones have a hardware constraint that the texture size must be a power of 2 and at most $4096\times4096$, we split the large texture images into multiple $4096\times4096$ texture images.

\section{Quadrature details}
\label{sec:supp_quadrature}
The regular-grid mesh $\Mesh$ provides an efficient way for computing intersections between a ray and the mesh of size $P \times P \times P$ in $O(P)$ complexity, as shown in \Figure{quadrature}.

First, we compute the set of voxels that are intersected by the ray. This involves solving $3P$ ray-plane intersections and using those intersection points to obtain at most $3P$ intersected voxels. This step is shown in \Figure{quadrature}a and Eq.~\ref{eq:ray_voxel_intersection}.

Then, we use the acceleration grid $\Grid \in \real^{P \times P \times P}$ to prune voxels that are unlikely to contain geometry, with respect to a threshold $\tau_\Grid=0.1$. This step is shown in \Figure{quadrature}b and Eq.~\ref{eq:acceleration}.

Finally, we compute intersections between the ray and the faces of $\Mesh$ that are incident to the voxel's vertex to obtain the final set of quadrature points. This step is shown in \Figure{quadrature}c and Eq.~\ref{eq:ray_meshface_intersection}.

During the first quarter of the training iterations, $\Grid$ may not be accurate, therefore we will keep all $3P$ intersected voxels regardless of $\tau_\Grid$, and keep $3P$ intersection points (Recall that if the mesh grid is a regular grid, there are at most $3P$ intersections). Then in the next quarter, we will use $\Grid$ to remove empty voxels and keep at most $3P/2$ non-empty voxels and $3P/2$ intersection points that are closest to the camera. In the rest of the training, we will keep $3P/4$. We also double the training batch size each time we halve the number of intersections.

For the concentric boxes in unbounded $360^{\circ}$ scenes, we will compute their intersections and keep all of them.

\section{Initial meshes}
\label{sec:supp_init_meshes}
In this section we detail the polygonal meshes used for synthetic $360^{\circ}$, forward-facing, and unbounded $360^{\circ}$ scenes, see Fig.~\ref{fig:init_mesh} for 2D illustrations.

We will call the coordinate system of a regular mesh grid in a unit cube centered at the origin as the normalized coordinates, and we can apply transformations to obtain the grids in the world coordinates for different types of scenes.
In the following, we will denote points in the normalized coordinates as $\mathbf{p} \in [-0.5,0.5]$ and points in the world coordinates as $\mathbf{p}'$.

For synthetic $360^{\circ}$ scenes, we apply scaling to the grid to put the object inside the grid.
\begin{align}
\mathbf{p}' = w\mathbf{p},
\end{align}
where $w=2.4$ or $3$, depending on the size of the object. We use a grid size of $P=128$.

In forward-facing scenes, we apply transformation to concentrate more voxels close to the camera, as shown in Fig.~\ref{fig:init_mesh} (b).
\begin{align}
\begin{cases}
\mathbf{p}_z' &= \exp( w (\mathbf{p}_z + 0.5) ), \\
\mathbf{p}_x' &= u \mathbf{p}_x \mathbf{p}_z', \\
\mathbf{p}_y' &= v \mathbf{p}_y \mathbf{p}_z',
\end{cases}
\end{align}
where $w$ is set to a value so that $\mathbf{p}_z'=25$ when $\mathbf{p}_z=0.5$; $u=v=1.75$. We use a grid size of $P=128$.

In unbounded $360^{\circ}$ scenes, we assume the cameras are inside the unit cube in the normalized coordinates, therefore we do not apply transformations. However, to model the surrounding environments, we add a set of $L+1$ concentric boxes around the regular grid. The boxes have fixed positions and geometry, and their distances to the center are given by
\begin{align}
d_i = ( \exp(\frac{wi}{L} ) + w-1 )/2w,
\end{align}
where $i$ ranges from $0$ to $L$. $w$ is set to a value so that $d_{L}=8$, therefore $d_i \in [0.5,8]$. We use a grid size of $P=128$, and $L=64$.

\section{Per-Scene metrics}
\label{sec:supp_per_scene}

We provide per-scene breakdown for the quality metrics in Table~\ref{table:supp_psnr_syn}~\ref{table:supp_ssim_syn}~\ref{table:supp_lpips_syn}~\ref{table:supp_psnr_ff}~\ref{table:supp_ssim_ff}~\ref{table:supp_lpips_ff}~\ref{table:supp_psnr_r360}~\ref{table:supp_ssim_r360}~\ref{table:supp_lpips_r360}.
We provide per-scene breakdown for rendering speed and storage cost in Table~\ref{table:supp_fps_syn}~\ref{table:supp_fps_ff}~\ref{table:supp_fps_r360}, where OOM (out-of-memory) indicates the device cannot run a testing scene due to GPU memory issues, and ICP (incompatible) indicates the device cannot run the method due to compatibility issues. The GPU memory and disk storage were tested on the Desktop.

For Surface Pro 6, Gaming laptop, and Desktop, we disable frame-rate limiting from vertical synchronization by starting the Chrome browser with the following arguments:
\begin{verbatim}
--disable-frame-rate-limit
--disable-gpu-vsync
\end{verbatim}
However, for phones and Chromebook, we did not find a way to easily disable vertical synchronization, therefore the FPS is capped at 60.

\clearpage

\begin{table*}[!t]
\begin{center}

\begin{tabular}{lccccccccc}
\hline
& Chair & Drums & Ficus & Hotdog & Lego & Materials & Mic & Ship & Mean \\
\hline
NeRF~\cite{mildenhall2020nerf} & 33.00 & 25.01 & 30.13 & 36.18 & 32.54 & 29.62 & 32.91 & 28.65 & 31.00 \\
JAXNeRF~\cite{jaxnerf2020github} & 33.88 & 25.08 & 30.51 & 36.91 & 33.24 & 30.03 & 34.52 & 29.07 & 31.65 \\
SNeRG~\cite{hedman2021snerg} & 33.24 & 24.57 & 29.32 & 34.33 & 33.82 & 27.21 & 32.60 & 27.97 & 30.38 \\
Ours & 34.09 & 25.02 & 30.20 & 35.46 & 34.18 & 26.72 & 32.48 & 29.06 & 30.90 \\
\hline
\end{tabular}

\end{center}
\caption{
\textbf{PSNR$\uparrow$} on \textbf{Synthetic $360^{\circ}$ scenes}.
}
\label{table:supp_psnr_syn}
\end{table*}

\begin{table*}[!t]
\begin{center}

\begin{tabular}{lccccccccc}
\hline
& Chair & Drums & Ficus & Hotdog & Lego & Materials & Mic & Ship & Mean \\
\hline
NeRF~\cite{mildenhall2020nerf} & 0.967 & 0.925 & 0.964 & 0.974 & 0.961 & 0.949 & 0.980 & 0.856 & 0.947 \\
JAXNeRF~\cite{jaxnerf2020github} & 0.974 & 0.927 & 0.967 & 0.979 & 0.968 & 0.952 & 0.987 & 0.865 & 0.952 \\
SNeRG~\cite{hedman2021snerg} & 0.975 & 0.929 & 0.967 & 0.971 & 0.973 & 0.938 & 0.982 & 0.865 & 0.950 \\
Ours & 0.978 & 0.927 & 0.965 & 0.973 & 0.975 & 0.913 & 0.979 & 0.867 & 0.947 \\
\hline
\end{tabular}

\end{center}
\caption{
\textbf{SSIM$\uparrow$} on \textbf{Synthetic $360^{\circ}$ scenes}.
}
\label{table:supp_ssim_syn}
\end{table*}

\begin{table*}[!t]
\begin{center}

\begin{tabular}{lccccccccc}
\hline
& Chair & Drums & Ficus & Hotdog & Lego & Materials & Mic & Ship & Mean \\
\hline
NeRF~\cite{mildenhall2020nerf} & 0.046 & 0.091 & 0.044 & 0.121 & 0.050 & 0.063 & 0.028 & 0.206 & 0.081 \\
JAXNeRF~\cite{jaxnerf2020github} & 0.027 & 0.070 & 0.033 & 0.030 & 0.030 & 0.048 & 0.013 & 0.156 & 0.051 \\
SNeRG~\cite{hedman2021snerg} & 0.025 & 0.061 & 0.028 & 0.043 & 0.022 & 0.052 & 0.016 & 0.156 & 0.050 \\
Ours & 0.025 & 0.077 & 0.048 & 0.050 & 0.025 & 0.092 & 0.032 & 0.145 & 0.062 \\
\hline
\end{tabular}

\end{center}
\caption{
\textbf{LPIPS$\downarrow$} on \textbf{Synthetic $360^{\circ}$ scenes}.
}
\label{table:supp_lpips_syn}
\end{table*}

\begin{table*}[!t]
\begin{center}

\begin{tabular}{lccccccccc}
\hline
\multicolumn{10}{c}{SNeRG~\cite{hedman2021snerg}} \\
& Chair & Drums & Ficus & Hotdog & Lego & Materials & Mic & Ship & Mean \\
\hline
iPhone XS & OOM & OOM & OOM & OOM & OOM & OOM & OOM & OOM & - \\
Pixel 3 & OOM & OOM & OOM & OOM & OOM & OOM & OOM & OOM & - \\
Surface Pro 6 & ICP & ICP & ICP & ICP & ICP & ICP & ICP & ICP & - \\
Chromebook & 28.06 & OOM & OOM & 26.11 & 27.08 & 16.48 & 26.99 & 11.01 & 22.62 \\
Gaming laptop & 4.94 & 10.27 & OOM & 8.10 & 9.41 & 2.05 & 21.65 & 1.69 & 8.30 \\
Gaming laptop \pluggedin & 37.66 & 51.06 & OOM & 45.52 & 60.20 & 13.81 & 87.67 & 11.17 & 43.87 \\
Desktop \pluggedin & 120.70 & 147.72 & 81.88 & 436.05 & 232.03 & 92.45 & 507.54 & 39.73 & 207.26 \\
GPU memory & 1254.00 & 4729.00 & 8243.00 & 1253.00 & 1253.00 & 1253.00 & 1251.00 & 2422.00 & 2707.25 \\
Disk storage & 141.00 & 44.00 & 43.00 & 67.00 & 114.00 & 134.00 & 22.00 & 129.00 & 86.75 \\
\hline
 \\
\hline
\multicolumn{10}{c}{Ours} \\
& Chair & Drums & Ficus & Hotdog & Lego & Materials & Mic & Ship & Mean \\
\hline
iPhone XS & 60.00 & 60.00 & 60.00 & 60.00 & 50.10 & 54.65 & 60.00 & 42.37 & 55.89 \\
Pixel 3 & 41.68 & 38.71 & 43.09 & 35.59 & 29.56 & 32.35 & 52.65 & 23.52 & 37.14 \\
Surface Pro 6 & 83.40 & 83.15 & 99.34 & 64.01 & 57.11 & 58.80 & 130.62 & 42.76 & 77.40 \\
Chromebook & 60.00 & 60.00 & 60.00 & 53.24 & 47.51 & 51.04 & 60.00 & 37.56 & 53.67 \\
Gaming laptop & 186.03 & 183.04 & 231.01 & 156.08 & 118.27 & 129.80 & 332.10 & 89.74 & 178.26 \\
Gaming laptop \pluggedin & 657.77 & 656.22 & 643.32 & 649.58 & 566.39 & 618.98 & 648.88 & 412.70 & 606.73 \\
Desktop \pluggedin & 810.99 & 789.30 & 882.23 & 707.27 & 629.70 & 659.95 & 970.35 & 509.48 & 744.91 \\
GPU memory & 451.00 & 590.00 & 450.00 & 456.00 & 723.00 & 721.00 & 322.00 & 594.00 & 538.38 \\
Disk storage & 107.00 & 120.00 & 80.00 & 88.00 & 199.00 & 191.00 & 50.00 & 171.00 & 125.75 \\
\hline
\end{tabular}

\end{center}
\caption{
\textbf{Rendering speed} in frames per second (FPS), and \textbf{GPU memory and disk storage} in MB, on \textbf{Synthetic $360^{\circ}$ scenes}.
}
\label{table:supp_fps_syn}
\end{table*}

\clearpage

\begin{table*}[!t]
\begin{center}

\begin{tabular}{lccccccccc}
\hline
& Room & Fern & Leaves & Fortress & Orchids & Flower & Trex & Horns & Mean \\
\hline
NeRF~\cite{mildenhall2020nerf} & 32.70 & 25.17 & 20.92 & 31.16 & 20.36 & 27.40 & 26.80 & 27.45 & 26.50 \\
JAXNeRF~\cite{jaxnerf2020github} & 33.30 & 24.92 & 21.24 & 31.78 & 20.32 & 28.09 & 27.43 & 28.29 & 26.92 \\
SNeRG~\cite{hedman2021snerg} & 30.04 & 24.85 & 20.01 & 30.91 & 19.73 & 27.00 & 25.80 & 26.71 & 25.63 \\
Ours & 31.28 & 24.59 & 20.54 & 30.82 & 19.66 & 27.05 & 26.26 & 27.09 & 25.91 \\
\hline
\end{tabular}

\end{center}
\caption{
\textbf{PSNR$\uparrow$} on \textbf{Forward-facing scenes}.
}
\label{table:supp_psnr_ff}
\end{table*}

\begin{table*}[!t]
\begin{center}

\begin{tabular}{lccccccccc}
\hline
& Room & Fern & Leaves & Fortress & Orchids & Flower & Trex & Horns & Mean \\
\hline
NeRF~\cite{mildenhall2020nerf} & 0.948 & 0.792 & 0.690 & 0.881 & 0.641 & 0.827 & 0.880 & 0.828 & 0.811 \\
JAXNeRF~\cite{jaxnerf2020github} & 0.958 & 0.806 & 0.717 & 0.897 & 0.657 & 0.850 & 0.902 & 0.863 & 0.831 \\
SNeRG~\cite{hedman2021snerg} & 0.936 & 0.802 & 0.696 & 0.889 & 0.655 & 0.835 & 0.882 & 0.852 & 0.818 \\
Ours & 0.943 & 0.808 & 0.711 & 0.891 & 0.647 & 0.839 & 0.900 & 0.864 & 0.825 \\
\hline
\end{tabular}

\end{center}
\caption{
\textbf{SSIM$\uparrow$} on \textbf{Forward-facing scenes}.
}
\label{table:supp_ssim_ff}
\end{table*}

\begin{table*}[!t]
\begin{center}

\begin{tabular}{lccccccccc}
\hline
& Room & Fern & Leaves & Fortress & Orchids & Flower & Trex & Horns & Mean \\
\hline
NeRF~\cite{mildenhall2020nerf} & 0.178 & 0.280 & 0.316 & 0.171 & 0.321 & 0.219 & 0.249 & 0.268 & 0.250 \\
JAXNeRF~\cite{jaxnerf2020github} & 0.086 & 0.207 & 0.247 & 0.108 & 0.266 & 0.156 & 0.143 & 0.173 & 0.173 \\
SNeRG~\cite{hedman2021snerg} & 0.133 & 0.198 & 0.252 & 0.125 & 0.255 & 0.167 & 0.157 & 0.176 & 0.183 \\
Ours & 0.143 & 0.202 & 0.245 & 0.115 & 0.277 & 0.163 & 0.147 & 0.169 & 0.183 \\
\hline
\end{tabular}

\end{center}
\caption{
\textbf{LPIPS$\downarrow$} on \textbf{Forward-facing scenes}.
}
\label{table:supp_lpips_ff}
\end{table*}

\begin{table*}[!t]
\begin{center}

\begin{tabular}{lccccccccc}
\hline
\multicolumn{10}{c}{SNeRG~\cite{hedman2021snerg}} \\
& Room & Fern & Leaves & Fortress & Orchids & Flower & Trex & Horns & Mean \\
\hline
iPhone XS & OOM & OOM & OOM & OOM & OOM & OOM & OOM & OOM & - \\
Pixel 3 & OOM & OOM & OOM & OOM & OOM & OOM & OOM & OOM & - \\
Surface Pro 6 & ICP & ICP & ICP & ICP & ICP & ICP & ICP & ICP & - \\
Chromebook & 9.75 & 6.02 & OOM & 9.68 & OOM & 5.12 & 8.68 & OOM & 7.85 \\
Gaming laptop & 7.77 & 1.28 & 0.80 & 8.46 & 1.14 & 0.67 & 4.72 & 4.18 & 3.63 \\
Gaming laptop \pluggedin & 52.40 & 14.45 & 6.15 & 54.47 & 12.43 & 8.77 & 32.87 & 26.51 & 26.01 \\
Desktop \pluggedin & 110.36 & 28.18 & 13.54 & 122.91 & 17.59 & 15.96 & 62.65 & 34.46 & 50.71 \\
GPU memory & 3594.00 & 3585.00 & 4729.00 & 3595.00 & 5903.00 & 3593.00 & 3595.00 & 5903.00 & 4312.13 \\
Disk storage & 149.00 & 288.00 & 408.00 & 162.00 & 704.00 & 321.00 & 251.00 & 415.00 & 337.25 \\
\hline
 \\
\hline
\multicolumn{10}{c}{Ours} \\
& Room & Fern & Leaves & Fortress & Orchids & Flower & Trex & Horns & Mean \\
\hline
iPhone XS & 29.82 & 25.10 & OOM & 30.02 & OOM & 26.28 & 26.30 & 25.59 & 27.19 \\
Pixel 3 & 13.57 & 12.74 & 8.66 & 14.69 & 10.77 & 12.98 & 13.07 & 12.71 & 12.40 \\
Surface Pro 6 & 22.92 & 20.32 & 13.84 & 29.13 & 17.10 & 22.30 & 22.53 & 23.92 & 21.51 \\
Chromebook & 20.70 & 18.95 & 14.65 & 23.16 & 16.79 & 20.06 & 20.08 & 21.12 & 19.44 \\
Gaming laptop & 64.27 & 55.88 & 37.11 & 76.29 & 48.72 & 60.60 & 59.65 & 59.26 & 57.72 \\
Gaming laptop \pluggedin & 281.01 & 252.70 & 170.66 & 303.77 & 222.54 & 260.44 & 258.45 & 251.82 & 250.17 \\
Desktop \pluggedin & 377.87 & 352.01 & 254.51 & 397.00 & 323.54 & 367.68 & 359.68 & 362.46 & 349.34 \\
GPU memory & 610.00 & 610.00 & 1143.00 & 473.00 & 1276.00 & 611.00 & 604.00 & 747.00 & 759.25 \\
Disk storage & 127.00 & 147.00 & 353.00 & 89.00 & 372.00 & 151.00 & 162.00 & 211.00 & 201.50 \\
\hline
\end{tabular}

\end{center}
\caption{
\textbf{Rendering speed} in frames per second (FPS), and \textbf{GPU memory and disk storage} in MB, on \textbf{Forward-facing scenes}.
}
\label{table:supp_fps_ff}
\end{table*}

\clearpage

\begin{table*}[!t]
\begin{center}

\begin{tabular}{lcccccc}
\hline
& Bicycle & Flower  & Garden  & Stump & Treehill & Mean \\
\hline
JAXNeRF~\cite{jaxnerf2020github} & 21.76 & 19.40 & 23.11 & 21.73 & 21.28 & 21.46 \\
NeRF++~\cite{nerf++} & 22.64 & 20.31 & 24.32 & 24.34 & 22.20 & 22.76 \\
Ours & 21.70 & 18.86 & 23.54 & 23.95 & 21.72 & 21.95 \\
\hline
\end{tabular}

\end{center}
\caption{
\textbf{PSNR$\uparrow$} on \textbf{Unbounded $360^{\circ}$ scenes}.
}
\label{table:supp_psnr_r360}
\end{table*}

\begin{table*}[!t]
\begin{center}

\begin{tabular}{lcccccc}
\hline
& Bicycle & Flower  & Garden  & Stump & Treehill & Mean \\
\hline
JAXNeRF~\cite{jaxnerf2020github} & 0.455 & 0.376 & 0.546 & 0.453 & 0.459 & 0.458 \\
NeRF++~\cite{nerf++} & 0.526 & 0.453 & 0.635 & 0.594 & 0.530 & 0.548 \\
Ours & 0.426 & 0.321 & 0.599 & 0.556 & 0.450 & 0.470 \\
\hline
\end{tabular}

\end{center}
\caption{
\textbf{SSIM$\uparrow$} on \textbf{Unbounded $360^{\circ}$ scenes}.
}
\label{table:supp_ssim_r360}
\end{table*}

\begin{table*}[!t]
\begin{center}

\begin{tabular}{lcccccc}
\hline
& Bicycle & Flower  & Garden  & Stump & Treehill & Mean \\
\hline
JAXNeRF~\cite{jaxnerf2020github} & 0.536 & 0.529 & 0.415 & 0.551 & 0.546 & 0.515 \\
NeRF++~\cite{nerf++} & 0.455 & 0.466 & 0.331 & 0.416 & 0.466 & 0.427 \\
Ours & 0.513 & 0.526 & 0.358 & 0.430 & 0.522 & 0.470 \\
\hline
\end{tabular}

\end{center}
\caption{
\textbf{LPIPS$\downarrow$} on \textbf{Unbounded $360^{\circ}$ scenes}.
}
\label{table:supp_lpips_r360}
\end{table*}

\begin{table*}[!t]
\begin{center}

\begin{tabular}{lcccccc}
\hline
\multicolumn{7}{c}{Ours} \\
& Bicycle & Flower  & Garden  & Stump & Treehill & Mean \\
\hline
iPhone XS & OOM & OOM & 22.20 & OOM & OOM & 22.20 \\
Pixel 3 & 9.44 & 8.61 & 10.49 & 8.54 & 9.12 & 9.24 \\
Surface Pro 6 & 20.24 & 19.12 & 21.67 & 18.21 & 17.97 & 19.44 \\
Chromebook & 15.89 & 14.72 & 16.56 & 14.23 & 15.02 & 15.28 \\
Gaming laptop & 55.62 & 59.18 & 58.19 & 51.73 & 51.89 & 55.32 \\
Gaming laptop \pluggedin & 195.63 & 194.66 & 204.31 & 178.89 & 189.46 & 192.59 \\
Desktop \pluggedin & 280.24 & 282.02 & 295.74 & 265.90 & 274.58 & 279.70 \\
GPU memory & 1350.00 & 1081.00 & 808.00 & 1082.00 & 1490.00 & 1162.20 \\
Disk storage & 400.00 & 294.00 & 239.00 & 337.00 & 453.00 & 344.60 \\
\hline
\end{tabular}

\end{center}
\caption{
\textbf{Rendering speed} in frames per second (FPS), and \textbf{GPU memory and disk storage} in MB, on \textbf{Unbounded $360^{\circ}$ scenes}.
}
\label{table:supp_fps_r360}
\end{table*}

\clearpage

\end{document}